\documentclass[preprint,12pt]{elsarticle}

\usepackage{amssymb}
\usepackage{amsmath}
\usepackage{algorithm}
\usepackage{algorithmic}
\usepackage{lipsum}
\usepackage{lineno}
\usepackage{pgfplots}
\usepackage{natbib}
\usepackage{amsmath,amsfonts}
\usepackage{algorithmic}
\usepackage{graphicx}
\usepackage{textcomp}
\usepackage{xcolor}
\usepackage{booktabs}
\usepackage{lipsum}
\usepackage{makecell}
\usepackage{multirow}
\usepackage{subcaption}
\usepackage[symbol]{footmisc}
\usepackage{longtable}
\usepackage{url}
\usepackage{nameref}
\usepackage{placeins}
\usepackage{lineno}
\usepackage{tcolorbox}
\usepackage{float}
\usepackage{tabularx}

\pgfplotsset{compat=1.18} 

\journal{Online Social Networks and Media}

\begin{document}

\begin{frontmatter}



\title{Assessing the Potential of Generative Agents in Crowdsourced Fact-Checking}


\author{Luigia Costabile$^a$, Gian Marco Orlando$^a$, Valerio La Gatta$^b$, \newline Vincenzo Moscato$^a$} 

\affiliation{organization={Department of Electrical Engineering and Information Technology (DIETI), University of Naples Federico II},
            addressline={Via Claudio 21},
            city={Naples},
            country={Italy}}

\affiliation{organization={Northwestern University, Department of Computer Science, McCormick School of Engineering and Applied Science},
            addressline={2233 Tech Dr},
            city={Evanston},
            postcode={60208},
            state={IL},
            country={United States}}

\begin{abstract}

The growing spread of online misinformation has created an urgent need for scalable, reliable fact-checking solutions. Crowdsourced fact-checking—where non-experts evaluate claim veracity—offers a cost-effective alternative to expert verification, despite concerns about variability in quality and bias. Encouraged by promising results in certain contexts, major platforms such as X (formerly Twitter), Facebook, and Instagram have begun shifting from centralized moderation to decentralized, crowd-based approaches.

In parallel, advances in Large Language Models (LLMs) have shown strong performance across core fact-checking tasks, including claim detection and evidence evaluation. However, their potential role in crowdsourced workflows remains unexplored. This paper investigates whether LLM-powered generative agents—autonomous entities that emulate human behavior and decision-making—can meaningfully contribute to fact-checking tasks traditionally reserved for human crowds.

Using the protocol of La Barbera et al. (2024) \cite{Crowd_Fact_Checking_Base_Paper}, we simulate crowds of generative agents with diverse demographic and ideological profiles. Agents retrieve evidence, assess claims along multiple quality dimensions, and issue final veracity judgments. Our results show that agent crowds outperform human crowds in truthfulness classification, exhibit higher internal consistency, and show reduced susceptibility to social and cognitive biases. Compared to humans, agents rely more systematically on informative criteria such as \emph{Accuracy}, \emph{Precision}, and \emph{Informativeness}, suggesting a more structured decision-making process. Overall, our findings highlight the potential of generative agents as scalable, consistent, and less biased contributors to crowd-based fact-checking systems.

\end{abstract}




\end{frontmatter}




\section{Introduction} \label{sec:Introduction}

In an era marked by the rapid dissemination of information, both accurate and misleading, fact-checking has become a critical tool to uphold the integrity of public discourse \cite{walter2020fact}. Traditional fact-checking approaches often rely on professional fact-checkers who assess the veracity of claims using rigorous, evidence-based methodologies \cite{graves2017anatomy}. However, given the sheer volume of information circulating online, these methods alone are not sufficient to meet the demand. In response, crowdsourced fact-checking \cite{allen2021scaling} has emerged as a scalable alternative that leverages the collective intelligence of non-expert contributors to evaluate the truthfulness of statements. Specifically, crowdsourced fact-checking relies on the premise that the \textit{``wisdom of the crowd"} can, when properly aggregated, yield reliable outcomes \cite{howe2006rise}. For this reason, many platforms employ structured workflows, such as questionnaires or rating systems, to guide contributors and enhance the consistency of their judgments. Notably, prior research \cite{allen2021scaling, zhao2023variety} has shown that human crowds can achieve levels of accuracy comparable to expert fact-checkers in certain contexts, particularly when diverse perspectives and systematic aggregation techniques are applied.

However, the effectiveness of crowdsourced fact-checking is not without limitations. Human evaluators, whether individually or in groups, are susceptible to biases, varying levels of expertise, and subjective interpretations. These factors can lead to inconsistencies in assessments, particularly when claims require nuanced understanding or domain-specific knowledge \cite{MSE_2, MSE_1}. Additionally, scaling crowdsourced systems poses challenges in maintaining quality control, as increased evaluator participation often amplifies variability in judgments. These limitations become evident on platforms like X (formerly Twitter), which have adopted crowdsourced fact-checking initiatives but often face criticism for inconsistent or incomplete outcomes \cite{prollochs2022community}. The increasing interest in such methods, as demonstrated by Meta’s plans to replace centralized fact-checking with a crowdsourced alternative\footnote{\url{https://www.cbsnews.com/news/meta-facebook-instagram-fact-checking-mark-zuckerberg/}}, underscores the relevance of exploring new ways to operationalize crowdsourced fact-checking at scale. 

Recent advancements in Large Language Models (LLMs) have introduced novel opportunities to fact-checking workflows \cite{vykopal2024generativelargelanguagemodels}. For example, LLMs have demonstrated strong performance on several tasks related to fact-checking \cite{sahitaj2025automatedfactcheckingrealworldclaims, 10.1007/978-3-031-71210-4_1}, including complex claim decomposition \cite{zhang2023towards}, evidence retrieval \cite{zhang2024reinforcement}, and veracity prediction \cite{papageorgiou2024survey}. A particularly compelling innovation within the LLM paradigm is the emergence of \textit{generative agents}—autonomous computational entities capable of reasoning, adapting, and simulating diverse human-like behaviors. These agents have been employed to replicate complex social phenomena, including information diffusion~\cite{GenAgents_S3}, network formation~\cite{ASONAM2024}, and influence dynamics~\cite{Social_Norms_GenAgents}. In the realm of fact-checking applications, \cite{FactAgent} shows how a single generative agent can engage in reflective reasoning throughout the verification process, leveraging both its internal knowledge and external resources. These include tools for retrieving information from online sources, as well as mechanisms for assessing the credibility of the domains from which news claims originate. This integration allows the agent to emulate expert-like judgment across multiple dimensions of fake news detection. Similarly, \cite{MUSE} designs an agentic framework where a generative agent corrects multimodal posts containing misinformation by retrieving relevant content from the web.

To the best of our knowledge, no prior work has investigated the integration of generative agents within a crowdsourced fact-checking process. Motivated by this gap, and in light of the strategic shifts of major technology companies towards crowdsourced approaches (e.g., \emph{Birdwatch} \cite{Birdwatch_Twitter}), this work investigates the potential of generative agents to serve as participants in crowdsourced fact-checking. 

\subsection*{Contributions of this work}

{\color{black}In this paper, we propose a framework that simulates the crowdsourced fact-checking process through the use of generative agents\footnote{{\color{black}Our objective is not to develop a state-of-the-art fact-checking system, but to explore the viability of generative agents as synthetic annotators for crowd-based fact-checking tasks.}}. The framework is designed to replicate the experimental protocol presented in \cite{Crowd_Fact_Checking_Base_Paper} to ensure comparability between agents and human crowds. In particular, each generative agent in our simulation is assigned a realistic profile based on the demographic and ideological data (e.g., gender, age) collected in \cite{Crowd_Fact_Checking_Base_Paper}. The verification process unfolds in two key stages. First, in the \textit{Evidence Selection} phase, generative agents choose the resource (i.e., summarised web page) to use as evidence for assessing the truthfulness of a given claim. Then, in the \textit{Questionnaire Completion} phase, agents were prompted to fill the structured questionnaire adopted in \cite{Crowd_Fact_Checking_Base_Paper}, which involves a multi-dimensional evaluation of each claim. Specifically, agents rate the statement across several dimensions—\emph{Accuracy}, \emph{Unbiasedness}, \emph{Comprehensibility}, \emph{Precision}, \emph{Completeness}, and \emph{Speaker's Trustworthiness}—before issuing a final veracity judgment.

We leverage this framework to replicate the human experiment performed in \cite{Crowd_Fact_Checking_Base_Paper}. Specifically, we selected 70 claims used for the original experiment and employed 50 agents, each tasked with fact-checking 14 distinct claims. In line with the original human experiment, we assigned claims to agents such that the same claim was reviewed by 10 agents. Eventually, we compare the results obtained from generative agents with the original annotations provided by humans, and answer the following research questions (RQs):

\begin{itemize}
    \item[\textbf{RQ1:}] \emph{How effective is fact-checking by generative agents compared to human crowds?} \\
    As LLMs are increasingly considered for tasks traditionally assigned to human annotators, it remains unclear whether they can deliver crowd-level accuracy and reliability in the context of misinformation detection. This question focuses on overall effectiveness, allowing us to assess whether generative agents can match or exceed the human crowd in aligning with expert ground truth.

    \item[\textbf{RQ2:}] \emph{What is the role of external evidence and claim recency in generative agents' fact-checking performance?} \\
    While LLMs encode extensive prior knowledge, real-world misinformation detection often hinges on timely access to current, context-specific evidence. This question examines whether generative agents are capable of leveraging external evidence and how this capability depends on the claim recency.

    \item[\textbf{RQ3:}] \emph{Do humans and generative agents rely on the same factors when assessing truthfulness?} \\
    Understanding the alignment (or misalignment) in evaluative strategies is critical to assessing the trustworthiness and interpretability of LLM-based judgments. By comparing reliance on claim-level characteristics (e.g., completeness, speaker's trustworthiness) and user attributes (e.g., demographics, ideology), we investigate whether generative agents can emulate both the diversity and subjectivity of human evaluators—or offer a more consistent alternative.
\end{itemize}


To summarize, the key contributions of this work are as follows:
\begin{itemize}
    \item We propose a framework that simulates the crowdsourced fact-checking process using generative agents with realistic demographic and ideological profiles;
    \item We replicate the experimental protocol of \cite{Crowd_Fact_Checking_Base_Paper} to enable a direct comparison between human and agent-based crowds powered by three different LLMs—Llama 3.1 8B, Gemma 2 9B, and Mistral 7B;
    \item We conduct a comprehensive evaluation across three core research questions, examining \textit{(i)} the effectiveness of generative agents compared to human crowds, \textit{(ii)} the impact of external evidence and claim recency on agent performance, and \textit{(iii)} the extent to which agents replicate human evaluative strategies and exhibit demographic or ideological biases.
\end{itemize}
}

This paper is organized as follows: \nameref{sec:Introduction} section outlines the context and motivation for this research. \nameref{sec:RW} section reviews relevant literature that informed our study. \nameref{sec:Methodology} section describes the datasets, agent design, and simulation workflow. \nameref{sec:Experiments} section presents the experiments conducted to address the RQs. {\color{black}The \nameref{sec:Discussion} section reflects on the broader implications of our findings and discusses key limitations.} Finally, \nameref{sec:Conclusion} section summarizes the findings and suggests directions for future research.

\section{Related Work} \label{sec:RW}

\subsection{Crowdsourced Fact-Checking}

Crowdsourced fact-checking refers to the practice of leveraging large groups of non-expert individuals to evaluate the veracity of information \cite{10.1145/3694980}. This approach capitalizes on the collective intelligence and diverse perspectives of ordinary users \cite{10.1145/3694980}, operationalizing the principle of the \textit{``wisdom of the crowd''} \cite{howe2006rise}. For this reason, crowdsourced fact-checking has gained increasing traction as a scalable and cost-effective alternative to professional fact-checking in the fight against misinformation~\cite{allen2021scaling,Crowd_Fact_Checking_Base_Paper}.

A growing body of research has investigated the viability and limitations of crowdsourced approaches for assessing the truthfulness of online content~\cite{allen2021scaling, sethi2017crowdsourcing, zhao2023variety}. While the crowd can produce high-quality judgments under certain conditions, the effectiveness of this approach is influenced by several factors. Notably, the complexity and domain-specificity of claims—such as those related to political discourse—can significantly affect the accuracy of crowd assessments~\cite{draws2021checklist, eickhoff2018cognitive, hube2019understanding}. Furthermore, cognitive and ideological biases among contributors have been shown to compromise the reliability and consistency of crowd-sourced judgments~\cite{MSE_2, MSE_1}. For instance, individuals are more likely to scrutinize content from ideological opponents while refraining from critically evaluating claims aligned with their own beliefs~\cite{allen2022birds}. In addition, prior user evaluations can influence subsequent judgments, potentially introducing herding effects~\cite{panizza2023online}.

Despite these challenges, real-world implementations of crowdsourced fact-checking have demonstrated considerable potential. A prominent example is the \textit{Community Notes} system on X (formerly Twitter), previously known as \textit{Birdwatch}, which enables users to collaboratively annotate and contextualize potentially misleading content. Empirical analyses indicate that crowd-generated notes often converge with expert assessments in identifying check-worthy information \cite{Birdwatch_Twitter} and, in some cases, outperform professional fact-checkers in terms of speed \cite{ prollochs2022community}. In addition, \cite{wojcik2022birdwatch} proposed a hybrid model that integrates crowd contributions with algorithmic curation, demonstrating that such an approach can effectively reduce users’ engagement with misinformation. Comparative evaluations further suggest that community-driven fact-checking and expert-based methods can serve complementary roles in the broader misinformation mitigation ecosystem \cite{pilarski2024community}.

{\color{black}Overall, prior work has primarily focused on evaluating the effectiveness, reliability, and scalability of crowdsourced fact-checking when performed by human annotators. Our work builds on this well-established research field but introduces a new way of \textit{constructing} the crowd: we simulate a crowd composed of generative agents powered by LLMs. These agents are instantiated to reflect the demographic and ideological heterogeneity observed among real human annotators, and are prompted accordingly—without any task-specific fine-tuning—to emulate the behavior of ordinary crowd workers. By operationalizing the \emph{wisdom of the crowd} through an ensemble of synthetic profiles, we investigate whether an artificial crowd can replicate—or potentially exceed—the performance of human participants in misinformation detection.
}


\subsection{Automatic Fact-Checking approaches}

A substantial body of research has emerged around automated fact-checking, aiming to provide scalable solutions to the growing challenge of online misinformation \cite{vlachos2014fact, thorne2018automated, pan2023fact}. These systems typically follow a structured pipeline that includes claim detection, evidence retrieval, and claim verification \cite{guo2022survey}, enabling partial automation of tasks traditionally performed by human fact-checkers \cite{chakraborty2023information}. Despite their efficiency, fully-automated methods often fall short in solving these task, especially in scenarios that require multi-hop inference \cite{hassan2015detecting}, the aggregation of evidence from diverse, unstructured sources \cite{adair2017progress}, or the generalization across different domains \cite{nenno2024checkworthiness}. In response to these challenges, hybrid frameworks that incorporate human-in-the-loop components have been proposed \cite{demartini2020human, qu2022combining}, offering a middle ground that combines the scalability of automatic verification with the contextual discernment of human evaluators.

Recent advancements in Large Language Models (LLMs) have propelled the development of more sophisticated approaches to automatic fact-checking \cite{sahitaj2025automatedfactcheckingrealworldclaims, 10.1007/978-3-031-71210-4_1}. Leveraging their generative and reasoning capabilities, these models have been applied across the entire fact-checking pipeline—from identifying check-worthy claims \cite{majer-snajder-2024-claim,10.1145/3589335.3651504}, to retrieving relevant evidence \cite{tan2023evidencebasedinterpretableopendomainfactchecking,kangur2025evigenerate}, and ultimately assessing claim veracity \cite{tang-etal-2024-minicheck,giarelis-etal-2024-unified}. When combined with retrieval-augmented systems (RAG), LLMs enhance fact-checking accuracy and explanation \cite{zhang2024reinforcement}.  Additionally, building on the adaptability of LLMs in few-shot learning scenarios, in-context learning approaches decompose complex claims into simpler subclaims, which are progressively verified through structured question-answering steps \cite{zhang2023towards}. The potential of LLMs for fact-checking is further highlighted by approaches such as MUSE, which augments LLMs with real-time access to up-to-date information and multimodal retrieval to address misinformation across diverse platforms \cite{MUSE}.

More recently, \emph{generative agents} -- autonomous entities powered by LLMs -- have emerged as a way to simulate human behavior, including reasoning, adaptation, and interaction and broader social dynamics \cite{ASONAM2024, WebSci2025, GenAgents_S3, Social_Norms_GenAgents}. To the best of our knowledge, \cite{FactAgent} represents the only notable attempt that employs generative agents into the fact-checking workflow. In this case,  individual agents operate in isolation to perform news verification using their internal capabilities (e.g., language analysis) and external tools (e.g., URL analyis, news retrieval). 

Despite these innovations, existing approaches have primarily focused on automating individual components of the fact-checking process. By contrast, crowdsourced fact-checking has remained largely dependent on human participation. Our work breaks new ground on the possibility to automate crowdsourced fact-checking using generative agents in place of human contributors. Specifically, we introduce a framework in which an ensemble of diverse generative agents collectively evaluates the veracity of claims, mimicking the decentralized, consensus-driven nature of human crowdsourcing. Our goal is not to introduce a new state-of-the-art fact-checking system, but rather to assess whether artificial crowds composed of generative agents can replicate or outperform human crowds in crowdsourced fact-checking. Thus, this contribution offers a hybrid perspective that bridges the scalability of automated verification with the nuanced deliberation characteristic of human-based fact-checking.


\section{Materials and Methods} \label{sec:Methodology}

\subsection{Dataset}

We utilize a publicly available dataset initially employed in a prior crowdsourced fact-checking study \cite{Crowd_Fact_Checking_Base_Paper}. This dataset originates from a human-subject experiment involving 200 participants, who evaluated a total of 122 political and social statements made between 2008 and 2022. The statements were annotated by expert fact-checkers using a 6-level scale: \textit{Pants-On-Fire}, \textit{False}, \textit{Mostly-False}, \textit{Half-True}, \textit{Mostly-True}, and \textit{True}. Further details about this scale are provided in the subsequent \nameref{par:Questionnaire_Completion} section.

For each statement, the dataset includes a collection of 10 URLs corresponding to web pages that participants in the original study identified as relevant for assessing the truthfulness of the respective statements. Additionally, the dataset contains metadata on participants, including demographic attributes and self-reported political preferences.



\subsection{Methodology}

Our framework emulates the crowd-based fact-checking process by leveraging generative agents designed to replicate the diversity and decision-making processes of human participants. These agents are equipped with realistic profiles and operate within a structured workflow to perform key fact-checking tasks.

\begin{figure*}[t]
    \centering
    \includegraphics[width=1\linewidth]{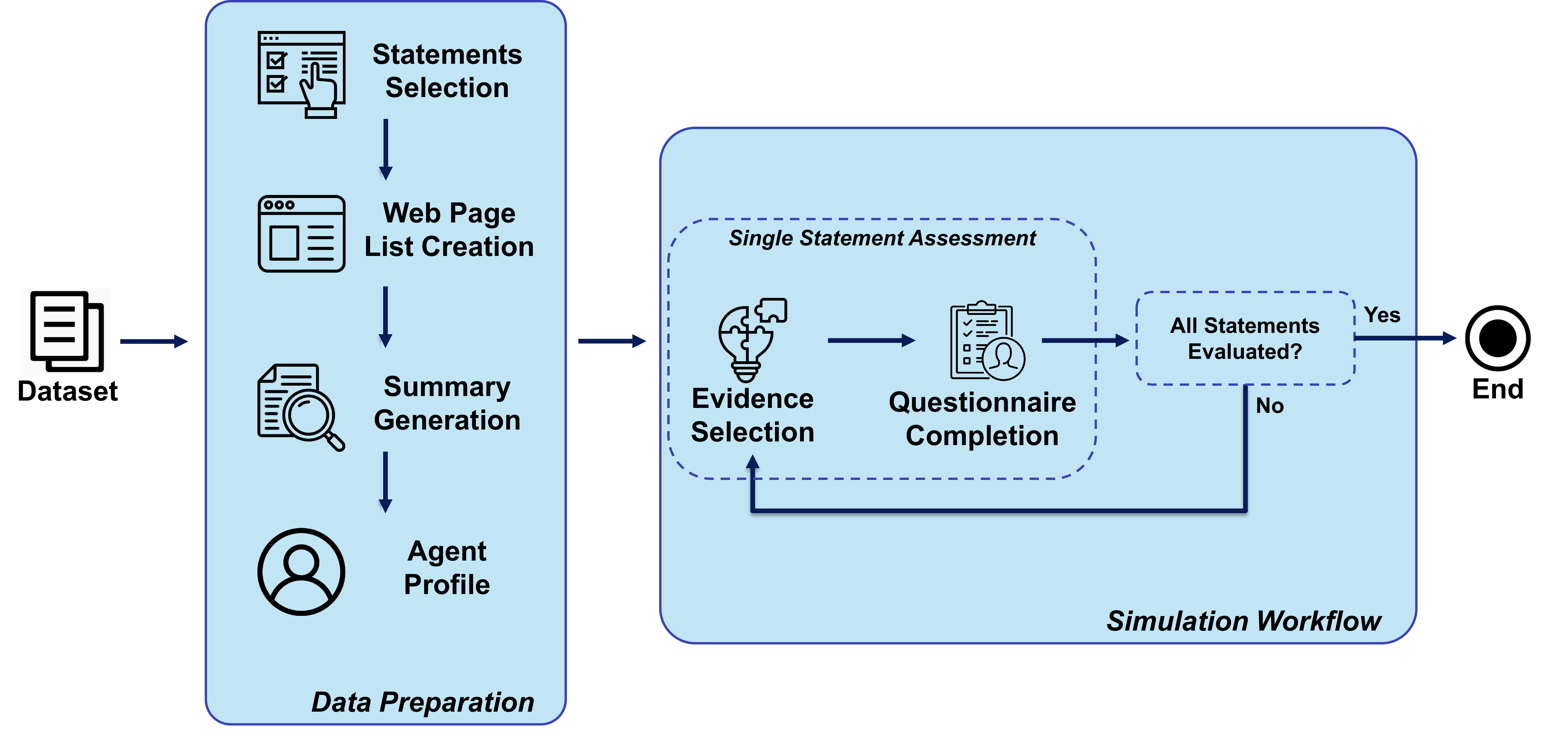}
    \caption{\textbf{Framework Overview.} The framework consists of two main phases: data preparation and simulation. During the data preparation phase, the dataset is tailored for generative agents, involving statement selection, evidence curation, and agent profile design to replicate human crowd diversity. The simulation phase mirrors a crowd-based fact-checking process, where generative agents perform two primary tasks: selecting the most relevant evidence and completing a structured questionnaire to assess the truthfulness and quality dimensions of statements.}
    \label{fig:Methodology}
\end{figure*}

The methodology unfolds in two main phases, as depicted in Figure \ref{fig:Methodology}: data preparation and simulation. During the data preparation phase, the dataset is tailored to align with the capabilities of generative agents. The simulation phase mirrors the fact-checking process by tasking agents to evaluate the truthfulness of statements through a structured questionnaire. Details of these processes are outlined in the following sections.

\subsubsection{Data Preparation} \label{sec:data_preparation}

The data preparation phase was critical for ensuring the reliability and relevance of the simulation. This process involved tailoring the original dataset to handle the transition from a human-based crowd to a system employing generative agents. The preparation encompassed several steps: selecting statements, curating web page data, generating summaries, and defining agent profiles.

\paragraph{\textit{Statements Selection}}

A subset of 70 statements from the original dataset was manually selected, focusing exclusively on claims made in 2022. This decision was motivated by the need to evaluate content more representative of the types of information commonly shared on contemporary social media platforms. Including older statements could introduce inaccuracies due to shifts in context, language usage, and public discourse over time. 
{\color{black} We investigate the impact of claim recency on generative agents' performance in our experiments.

To assess whether generative agents demonstrate topic-dependent variability in fact-checking performance, we categorized the 70 claims made in 2022 into three thematic areas using BERTopic \cite{grootendorst2022bertopic}. Specifically, BERTopic was employed to cluster the claims based on semantic similarity. Then, the authors of this manuscript manually annotated the topic of the claims in each cluster: \textit{Civil Rights} (17 claims), \textit{Conspiracy Theories} (25 claims), and \textit{Economics} (28 claims). Table \ref{tab:topic_examples} reports representative examples of fact-checked claims for each topic.}

\begin{table*}[t]
    \centering
    \caption{{\color{black}\textbf{Claim examples for each of the three topics identified.} The claims were grouped using BERTopic and subsequently manually annotated.}}
    \label{tab:topic_examples}
    \resizebox{\linewidth}{!}{%
    {\color{black}
    \begin{tabular}{lp{5cm}p{5cm}cc}
        \toprule
        \textbf{Topic} & \textbf{Claim} & \textbf{Speaker} & \textbf{Date} & \textbf{Ground Truth}\\
        \midrule
        \multirow{2}{*}{Civil Rights} 
            & When the New York State Senate voted to legalize abortion in 1970, 12 Republican senators voted in favor of it. & Andrea Stewart-Cousins & 03/05/2022 & True\\
        \cmidrule{2-5}
            & Gun manufacturers are the only industry in the country that have immunity from lawsuits. & Joe Biden & 02/06/2022 & False\\
        \midrule
        \multirow{2}{*}{Conspiracy Theories}
            & Cheri Beasley voted to reverse the conviction of an armed kidnapper, release a double murderer early. & National Republican Senatorial Committee & 14/06/2022 & True \\
        \cmidrule{2-5}
            & The 2021 Georgia Senate runoff and the 2020 presidential election were stolen. & David Perdue & 25/03/2022 & False\\
        \midrule
        \multirow{2}{*}{Economics}
            & The current spike in gas prices is largely the fault of Vladimir Putin. & Joe Biden & 14/03/2022 & True \\
        \cmidrule{2-5}
            & American oil is  more affordable than foreign oil. & Kevin McCarthy & 17/06/2022 & False \\
        \bottomrule
    \end{tabular}
    }
    }
\end{table*}

\paragraph{\textit{Web Page List Creation}}

Each statement was accompanied by a set of web pages serving as potential evidence for the fact-checking task. These web pages were systematically verified for accessibility. Only functional and accessible pages were retained, ensuring the inclusion of high-quality and reliable web resources in the analysis. Each retained web page featured a news article, providing generative agents the necessary context and evidence for evaluating the truthfulness of the associated statement.

\paragraph{\textit{Summary Generation}}

Incorporating the complete content of each news article within a single prompt presents practical challenges due to the limited context window of any LLM-powered agent and the increased risk of generating hallucinations \cite{LLM_Hallucinations}. To mitigate these challenges, a summary of the content was generated for each selected web page. These summaries are designed to retain key details relevant for assessing the truthfulness of each statement, providing generative agents with concise yet comprehensive evidence.

The summarization process employed a dual-step methodology. Initially, textual content was extracted from each web page using web scraping techniques. Subsequently, the extracted text was processed by a LLM to produce summaries guided by a carefully designed prompt. This prompt explicitly included the statement associated with each URL, ensuring the model to focus on extracting information relevant to the statement's context. This approach ensured the generated summaries were concise, contextually relevant, and aligned with the requirements of the fact-checking task.

{\color{black}To promote transparency and facilitate reproducibility, \ref{app:summarization_example} presents a detailed example of this process, including one claim and two associated web pages, along with the prompt used for summarization and the summaries generated from each source. The full set of claims, their associated evidence pages, and corresponding summaries are publicly available in the repository\footnote{\url{https://github.com/PRAISELab-PicusLab/generative-agents-crowdsourced-fact-checking}}.}

\paragraph{\textit{Agent Profile}}

The agent profiles were defined using the demographic and political data collected from the 200 participants in \cite{Crowd_Fact_Checking_Base_Paper}. Each agent's profile was designed to reflect key attributes such as age, gender, ethnicity, educational background, political alignment, and personal views on socio-political issues. This step aimed to replicate the diversity and characteristics of the human crowd by incorporating this wide range of demographic and ideological perspectives, ensuring that the decision-making processes of the agents closely mirrored those of their human counterparts. Further implementation details are provided in \ref{app:System_Prompt}.

\subsubsection{Simulation Workflow} \label{sec:simulation_workflow}

The simulation workflow consists of two phases designed to emulate the crowd fact-checking process using generative agents. These agents mimic human participants, performing two primary tasks: selecting the web page deemed most relevant as evidence for verifying a given statement and completing a questionnaire to evaluate the statement's truthfulness in the context of the fact-checking task.

\paragraph{\textit{Evidence Selection}} \label{par:Evidence_Selection}

In this phase, each agent selects the most relevant web page from the list provided for each statement. In line with previous studies \cite{Crowd_Fact_Checking_Base_Paper}, the agent is restricted to choose only one evidence judged as the most suitable source for verifying the truthfulness of the current statement. The agents are required to evaluate the sources based on their relevance, credibility, and the depth of information they offer regarding the statement. Additional details on how this selection process was operationalized are available in \ref{app:Main_Prompt}.

\paragraph{\textit{Questionnaire Completion}} \label{par:Questionnaire_Completion}

Agents perform the fact-checking task for a given statement by completing a structured questionnaire using a prompt-based approach. To enable a fair comparison with human-based fact-checking, the same questionnaire utilized in the prior study by \cite{Crowd_Fact_Checking_Base_Paper} is adopted. To support this process, agents are provided with the statement under evaluation and the summary of the web page selected during the \nameref{par:Evidence_Selection} phase. This ensures that the agent has all the necessary information to assess the statement's truthfulness and generate an accurate and informed response to the questionnaire. Each evaluation is accompanied by a justification to ensure the reasoning behind the evaluation is transparent and grounded in the evidence provided. Further details on the implementation are provided in \ref{app:Questionnaire_Prompt}.

In line with previous works \cite{Crowd_Fact_Checking_Base_Paper, Internal_Agreement_2}, statements were annotated by expert judges on a 6-level truthfulness scale which served as the ground truth:

\begin{itemize}
    \item \emph{Pants-On-Fire}, if the statement is not accurate or is not correct but also makes a ridiculous claim;
    \item \emph{False}, if the statement is not accurate or is not correct;
    \item \emph{Mostly-False}, if the statement contains an element of truth but ignores critical facts that would give a different impression;
    \item \emph{Half-True}, if the statement is partially accurate but leaves out important details or takes things out of context;
    \item \emph{Mostly-True}, if the statement is accurate but needs clarification or additional information;
    \item \emph{True}, if the statement is accurate and there’s nothing significant missing.
\end{itemize}

These values were then mapped to a 2-level scale, with the first three levels (\textit{Pants-On-Fire}, \textit{False}, \textit{Mostly-False}) classified as \textit{False}, and the last three levels (\textit{Half-True}, \textit{Mostly-True}, \textit{True}) classified as \textit{True}.

Additionally, the questionnaire asked the agents to evaluate the following quality dimensions for each statement:

\begin{itemize}
    \item \emph{Accuracy} assesses if the statement accurately reflects the topic without errors or incorrect information;
    \item \emph{Unbiasedness} determines if the statement is neutrally and objectively expressed, avoiding subjective or biased language;
    \item \emph{Comprehensibility} rates the statement’s clarity and readability, determining if it is easy to understand;
    \item \emph{Precision} evaluates whether the information in the statement is specific and detailed rather than vague or ambiguous;
    \item \emph{Completeness} assesses if the statement provides a full, comprehensive view of the topic, rather than only partial information;
    \item \emph{Speaker's Trustworthiness} rates the general trustworthiness of the speaker or author, based on reliability and credibility;
    \item \emph{Informativeness} judges if the statement provides valuable information or insights, rather than well-known facts or tautologies.
\end{itemize}

The values for these dimensions were scored on a 5-level scale: \textit{"Completely Disagree"}, \textit{"Partially Disagree"}, \textit{"Neutral"}, \textit{"Partially Agree"}, and \textit{"Completely Agree"}.

\section{Experiments} \label{sec:Experiments}

\subsection{Implementation Details}

We conducted a simulation involving 50 generative agents, each tasked with fact-checking 14 distinct statements. In total, 70 statements were annotated, with each statement independently reviewed by 10 agents, ensuring consistency with the crowd-based experimental design described in \cite{Crowd_Fact_Checking_Base_Paper}. Although the number of agents was reduced compared to the original study, the proportional distributions from the human crowd were preserved to maintain comparability. This design ensures that any differences in performance can be attributed solely to the intrinsic characteristics of generative agents versus humans, eliminating potential confounding variables and enabling a robust comparison under identical conditions. 
 Table \ref{tab:Implementation_Details} summarizes the composition of agents compared to the original human crowd, highlighting the alignment in distributions. {\color{black} The demographic and ideological categories, while not exhaustive (e.g., we do not model non-binary gender), replicate those in \cite{Crowd_Fact_Checking_Base_Paper} and allows a fair and direct comparison with the original human experiment.}

\begin{table*}[t]
    \centering
    \caption{\textbf{Demographic and ideological distributions of generative agents compared to the human crowd.} The percentage composition illustrates the alignment between generative agents and the original human crowd across key attributes.}
    \label{tab:Implementation_Details}
    \resizebox{\linewidth}{!}{%
    \begin{tabular}{llccc}
        \toprule
        \textbf{Trait} & \textbf{Category} 
        & \textbf{No. Agents} & \textbf{Agents Composition (\%)} & \textbf{Humans Composition (\%)}
        \\
        \midrule
        \multirow{2}{*}{\textbf{Ethnicity}} 
        & White & 34 & 68\% & 73.5\% \\
        & Black & 12 & 24\% & 15\% \\
        \midrule
        \multirow{3}{*}{\textbf{Political Faction}}
        & Democrats & 26 & 32\% & 32\% \\
        & Republicans & 13 & 17\% & 17\% \\
        & Independents & 11 & 22\% & 22.5\% \\
        \midrule
        \multirow{6}{*}{\textbf{Education Level}}
        & Post-graduate Degree & 9 & 18\% & 20.5\% \\
        & Post-graduate Schooling & 3 & 6\% & 4\% \\
        & Bachelor's Degree & 20 & 40\% & 39\% \\
        & College & 11 & 22\% & 22\% \\
        & High School & 6 & 12\% & 12.5\% \\
        & Less than High School & 1 & 2\% & 2\% \\
        \midrule
        \multirow{4}{*}{\textbf{Age}}
        & 19-25 & 5 & 10\% & 10\% \\
        & 26-35 & 15 & 30\% & 30\% \\
        & 36-50 & 18 & 36\% & 36\% \\
        & 51-80 & 12 & 24\% & 25\% \\
        \midrule
        \multirow{2}{*}{\textbf{Gender}}
        & Male & 30 & 60\% & 60\% \\
        & Female & 20 & 40\% & 40\% \\
        \bottomrule
    \end{tabular}
    }
\end{table*}

The framework\footnote{The code will be made available upon acceptance.} is implemented using PyAutogen \cite{AutoGen} to instantiate the agents. We repeated experiments using three different open-source LLMs: Llama 3.1 8B\footnote{https://ai.meta.com/blog/meta-llama-3-1/}, Gemma 2 9B\footnote{https://ai.google.dev/gemma}, and Mistral 7B\footnote{https://mistral.ai/news/announcing-mistral-7b/}. These models were selected to ensure a comprehensive evaluation across varying architectures and capabilities. Their selection is based on their unrestricted nature and data filtering policies designed to mitigate alignment and bias issues. To ensure full transparency and reproducibility of our methodology, we provide the complete prompts used in our experiments in the \ref{Prompt_Design_Appendix}.

For the \textit{Summary Generation} phase, textual content from each selected web page was extracted using the Python library BeautifulSoup\footnote{https://www.crummy.com/software/BeautifulSoup/}. The extracted content was then summarized using the GPT-4o as LLM via the OpenAI API\footnote{https://platform.openai.com/docs/models\#gpt-4o}.

\subsection{RQ1: How effective is fact-checking by generative agents compared to human crowd?}

We first focus on comparing the judgments of the crowd with the fact-checkers annotation. Subsequently, we assess the internal consistency of the evaluations within each group (humans or generative agents), providing an intrinsic measure of their coherence and reliability. {\color{black}In addition to the global performance analysis, we also investigate how the effectiveness of generative agents varies depending on claim topic and the number of agents evaluating each statement.}

\subsubsection{Comparison of Crowd Judgments with Expert Labels}

We evaluate the effectiveness of generative agents and human crowds with respect to fact-checkers. Specifically, we consider the average score assigned by the crowd and compared this score with the ground truth of each document. Table \ref{tab:RQ1_Results} shows accuracy, precision, recall and F1-score of generative agents and humans crowd on both 2-level and 6-level truthfulness scales. We observe that generative agents powered by Llama 3.1 always achieve better results than humans across all metrics, with improvements in the 6-level scale of 24\%, 34\%, 24\%, and 20\% in accuracy, precision, recall, and F1-score, respectively. {\color{black}Interestingly, Gemma 2 achieves the best recall score on the 2-level scale and the best precision score on the 6-level scale. Conversely, Mistral demonstrates the lowest performance across all metrics. In general, generative agents surpass human crowd for most metrics, particularly on the 2-level truthfulness scales.} These findings underscore the effectiveness of generative agents compared to human evaluators, particularly in binary classification scenarios.


While these metrics provide a straightforward measure of correctness by quantifying the proportion of exact matches between predictions and the ground truth, it does not account for the magnitude of disagreements in cases where predictions deviate. For example, smaller deviations (e.g., a prediction of 4 compared to a ground truth of 5) should be penalized less than substantial mismatches (e.g., 1 compared to 5), reflecting a more nuanced understanding of agreement. For this reason, in line with previous work \cite{Crowd_Fact_Checking_Base_Paper}, we measure the Krippendorff’s alpha ($\alpha$) \cite{krippendorff2011computing} coefficient to quantify the reliability of agreement among raters.

\begin{table*}[t]
    \centering
    \caption{\textbf{Performance metrics and agreement measures for generative agents and human evaluators on 6-level and 2-level truthfulness scales.} The results highlight generative agents' superior performance and consistency compared to human crowd, particularly in binary classification tasks (2-level scale), while also demonstrating nuanced performance variations in multi-level classifications (6-level scale).}
    \label{tab:RQ1_Results}
    \resizebox{\linewidth}{!}{%
    \begin{tabular}{lccccccccccc}
        \toprule
        \textbf{Scale} & \textbf{Crowd} & \textbf{Accuracy} & \textbf{Precision} & \textbf{Recall} & \textbf{F1} & \textbf{Ext. Agreement} & \multicolumn{2}{c}{\textbf{Pairwise Agreement}} & \textbf{Internal Agreement} \\
        \cmidrule(lr){8-9}
        & & \textbf{(Correct/Total)} & & & & \textbf{\boldmath$\alpha$} & \textbf{Exact} & \textbf{Directional} & \textbf{\boldmath$\alpha$} \\
        \midrule
        2-Level & \begin{tabular}{@{}c@{}}Llama 3.1 \\ Gemma 2 \\ Mistral \\ Humans\end{tabular} & 
        \begin{tabular}{@{}c@{}}\textbf{0.957 (67/70)} \\ 0.885 (62/70)\\ 0.900 (63/70) \\ 0.885 (62/70)\end{tabular} & 
        \begin{tabular}{@{}c@{}}\textbf{0.970} \\ 0.829 \\ 0.868 \\ 0.885\end{tabular} &
        \begin{tabular}{@{}c@{}}0.942 \\ \textbf{0.971} \\ 0.943 \\ 0.885\end{tabular} &
        \begin{tabular}{@{}c@{}}\textbf{0.956} \\ 0.895 \\ 0.904 \\ 0.885\end{tabular} &
        \begin{tabular}{@{}c@{}}\textbf{0.914} \\ 0.771 \\ 0.801 \\ 0.773\end{tabular} &
        \begin{tabular}{@{}c@{}}- \\ - \\ - \\ -\end{tabular} &
        \begin{tabular}{@{}c@{}}- \\ - \\ - \\ -\end{tabular} &
        \begin{tabular}{@{}c@{}}0.881 \\ \textbf{0.933} \\ 0.845 \\ 0.154\end{tabular} \\
        \midrule
        6-Level & \begin{tabular}{@{}c@{}}Llama 3.1 \\ Gemma 2 \\ Mistral \\ Humans\end{tabular} & 
        \begin{tabular}{@{}c@{}}\textbf{0.585 (41/70)} \\ 0.471 (33/70) \\ 0.428 (30/70) \\ 0.471 (33/70)\end{tabular} & 
        \begin{tabular}{@{}c@{}}0.619 \\ \textbf{0.626} \\ 0.399 \\ 0.459\end{tabular} &
        \begin{tabular}{@{}c@{}}\textbf{0.585} \\ 0.471 \\ 0.428 \\ 0.471\end{tabular} &
        \begin{tabular}{@{}c@{}}\textbf{0.548} \\ 0.433 \\ 0.370 \\ 0.458\end{tabular} &
        \begin{tabular}{@{}c@{}}0.759 \\ \textbf{0.820} \\ 0.745 \\ 0.793\end{tabular} &
        \begin{tabular}{@{}c@{}}0.331 \\ \textbf{0.335} \\ 0.285 \\ 0.172\end{tabular} &
        \begin{tabular}{@{}c@{}}0.714 \\ \textbf{0.748} \\ 0.690 \\ 0.543\end{tabular} &
        \begin{tabular}{@{}c@{}}0.867 \\ \textbf{0.945} \\ 0.907 \\ 0.258\end{tabular} \\
        \bottomrule
    \end{tabular}
    }
\end{table*}


\textbf{External agreement}, assessed through $\alpha$, thus measures not only the alignment between crowd-generated annotations and expert-provided ground truth but also the degree of consistency across multi-level classifications. High external agreement indicates that the crowd's annotations are sufficiently accurate to support or even replace expert labels.

To compute $\alpha$, separate reliability matrices were constructed for the human crowd and the generative agents. Each matrix consists of two rows: the first row represents the aggregated assessments from the crowd — human or generative agents — calculated using the mean, while the second row contains the expert-provided ground truth. Each column in the matrix corresponds to a specific statement, enabling a direct comparison of crowd annotations with expert labels.

{\color{black}
The results, detailed in Table \ref{tab:RQ1_Results}, reveal distinct patterns across the 2-level and 6-level scales. On the 2-level scale, generative agents achieved a higher or comparable external agreement ($\alpha = 0.914$, $\alpha = 0.771$, $\alpha = 0.801$ for Llama 3.1, Gemma 2 and Mistral, respectively) compared to the human crowd ($\alpha = 0.773$). In contrast, we observe that, on the 6-level classification scale,  Gemma 2 achieves the highest external agreement ($\alpha = 0.820$), surpassing the human crowd ($\alpha = 0.793$). Notably, crowds powered by Llama 3.1 and Mistral perform slightly worse ($\alpha = 0.759$ and $\alpha = 0.745$, respectively). These results confirm the strong alignment of generative agents with expert judgments, further validating their potential for crowdsourced fact-checking.
}

Complementing external agreement, which evaluates the global alignment of crowd annotations with expert-provided ground truth, \textbf{pairwise agreement} analyzes the consistency between pairs of statements \cite{Pairwise_Agreement}. To effectively capture the nuanced distinctions inherent in the 6-level scale, pairwise agreement was computed exclusively using this granular scale. This approach allows for a more sensitive assessment of the annotators' ability to discern and accurately reflect differences between statement pairs. For each pair of statements $S_1$ and $S_2$, pairwise agreement is assessed by comparing the differences in annotations with those in the ground truth. Specifically, the difference in annotators’ evaluations, $a(S_1) - a(S_2)$, is compared with the corresponding ground truth difference, $\text{gt}(S_1) - \text{gt}(S_2)$. Two types of agreement are defined:

\begin{itemize}
    \item \textit{Exact Agreement}: Occurs when the difference between the assessments of two items match exactly with the difference in the expert-provided ground truth labels ($a(S_1) - a(S_2) = \text{gt}(S_1) - \text{gt}(S_2)$);
    
    \item \textit{Directional Agreement}: Occurs when the signs of the differences align between annotations and ground truth, or when both differences are zero ($\text{sign}(a(S_1) - a(S_2)) = \text{sign}(\text{gt}(S_1) - \text{gt}(S_2))$).
\end{itemize}

Pairwise agreement, averaged across all statement pairs, provides a measure of overall consistency. As shown in Table \ref{tab:RQ1_Results}, generative agents consistently outperformed the human crowd in both exact and directional agreement. This indicates that agents are more adept at identifying relationships between evaluated statements, aligning more closely with expert evaluations.

\subsubsection{Internal Consistency of Crowd Judgments}

In a crowd-based fact-checking environment, different annotators may assign varying labels to the same document due to variations in interpretation, knowledge, or judgment. To quantitatively assess the consistency within the group - whether generative agents or human workers - we calculate the \textbf{internal agreement}. This metric reflects the coherence of the crowd. 
A high internal agreement indicates a strong level of concordance within the group, regardless of whether the assessments align with expert labels \cite{Internal_Agreement_1}.

Following a similar approach to the computation of external agreement, Krippendorff’s alpha ($\alpha$) was employed to compute internal agreement by applying it solely to the crowd’s annotations, without reference to expert ground truth \cite{Internal_Agreement_2}.

Results, summarized in Table \ref{tab:RQ1_Results}, reveal a clear contrast between the two groups. On the 2-level scale, generative agents exhibited a high internal agreement ($\alpha \ge 0.845$ for all models), indicating a strong coherence in their assessments. In contrast, the human crowd showed a much lower internal agreement ($\alpha = 0.154$), suggesting greater variability and inconsistency in their evaluations. A similar pattern emerges on the 6-level scale, where generative agents maintained a strong internal agreement ($\alpha \ge 0.867$), while the human crowd exhibited only moderate consistency ($\alpha = 0.258$). These findings underscore the robustness of generative agents in maintaining consistent evaluations. Conversely, the lower internal agreement observed among human annotators suggests greater variability in their interpretations and decision-making processes.

\subsubsection{Distribution of Ratings per Label}

\begin{figure*}[t]
    \centering
    \begin{subfigure}{0.32\textwidth}
        \centering
        \includegraphics[width=\textwidth]{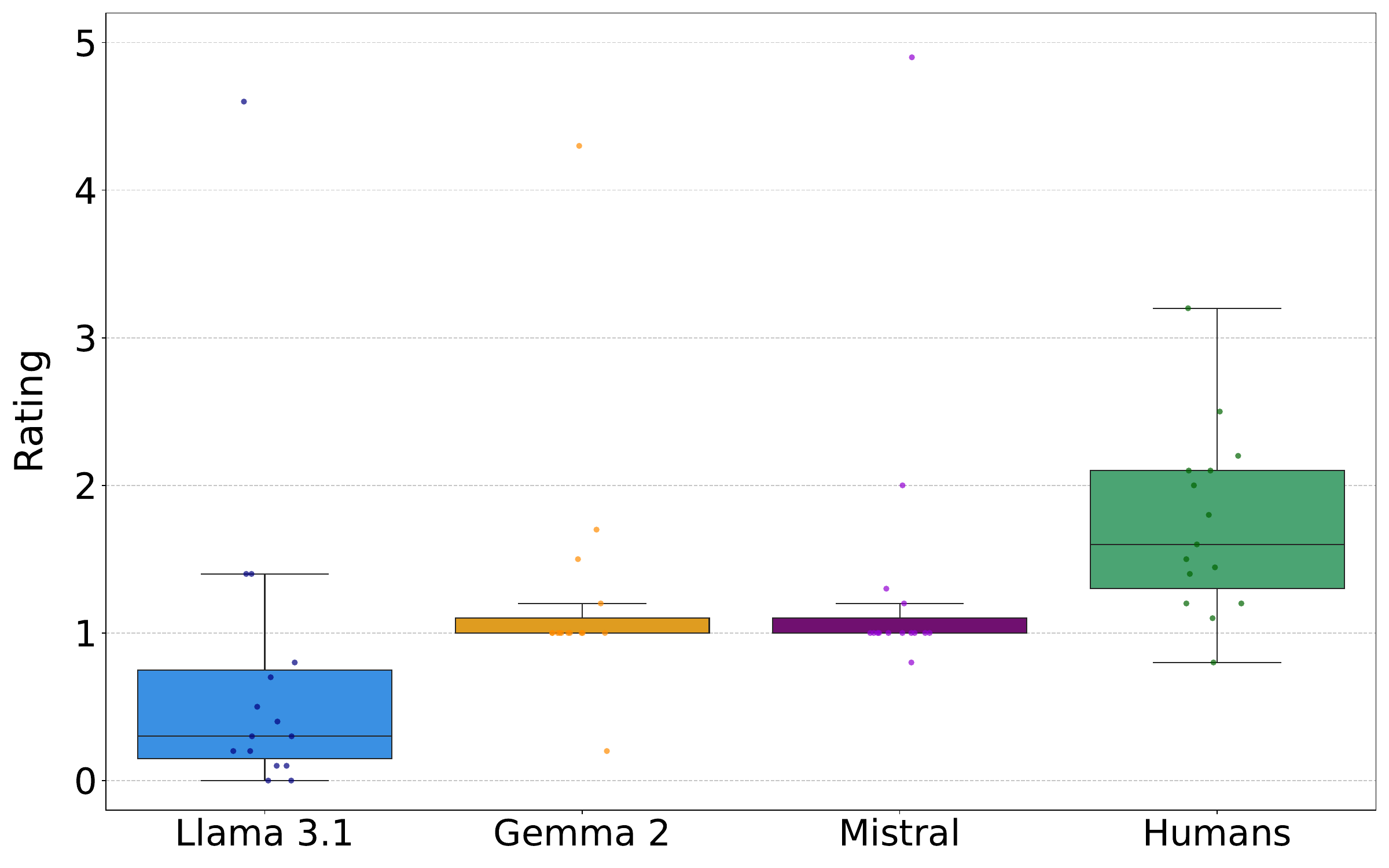}
        \caption{Pants-On-Fire} \label{fig:Distribution_Ratings_Pants_On_Fire}
    \end{subfigure}
    \begin{subfigure}{0.32\textwidth}
        \centering
        \includegraphics[width=\textwidth]{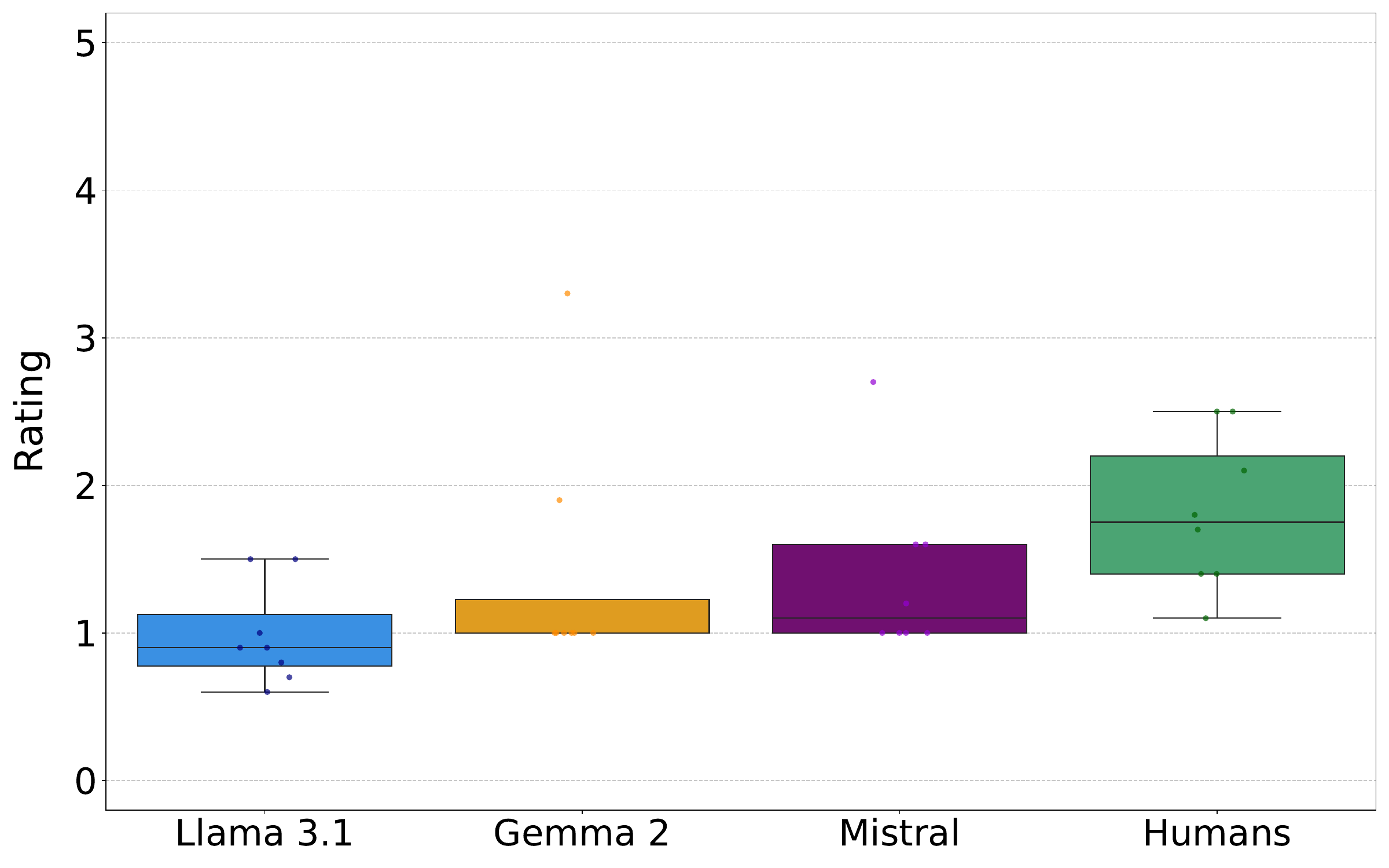}
        \caption{False} \label{fig:Distribution_Ratings_False}
    \end{subfigure}
    \begin{subfigure}{0.32\textwidth}
        \centering
        \includegraphics[width=\textwidth]{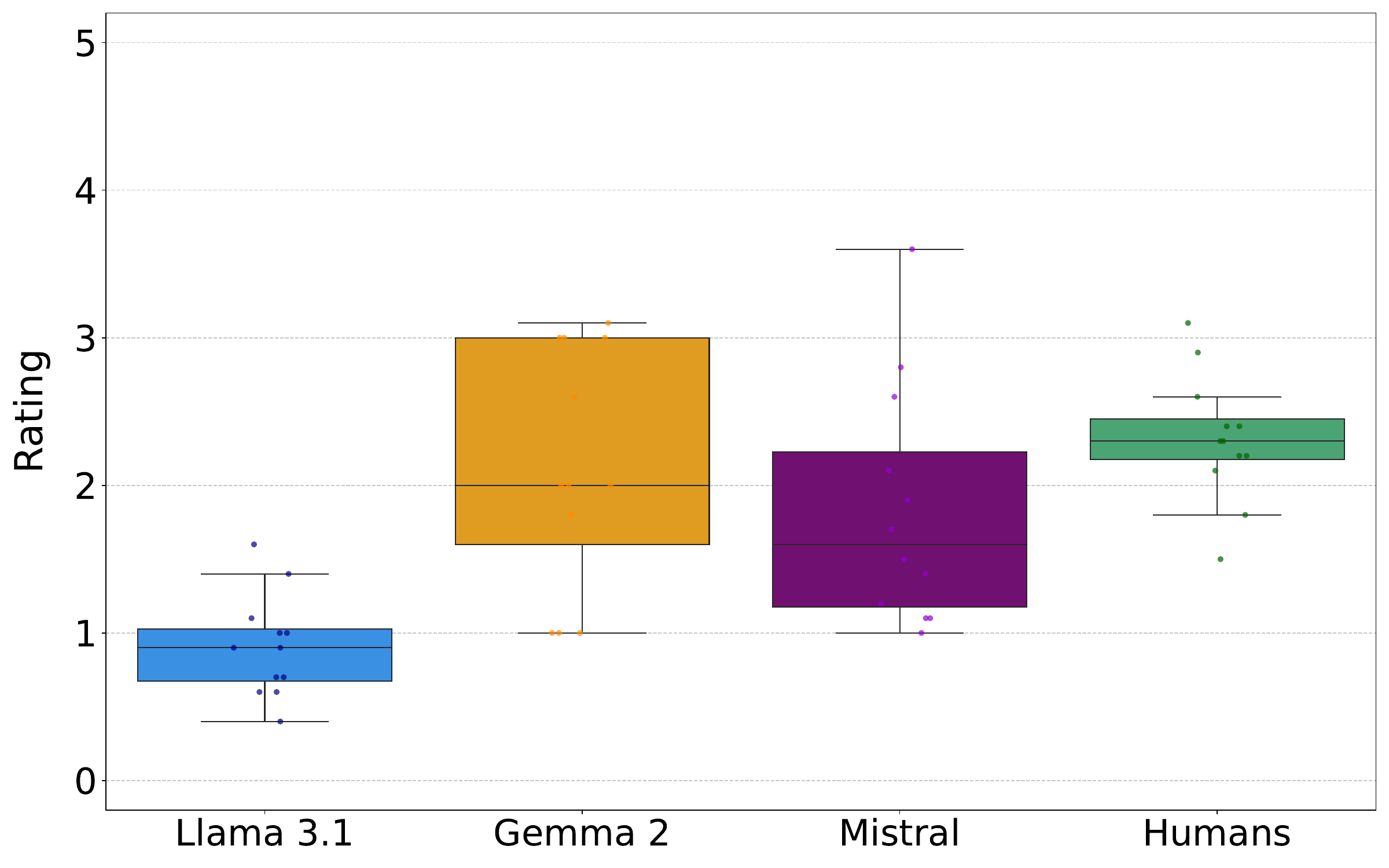}
        \caption{Mostly-False} \label{fig:Distribution_Ratings_Mostly_False}
    \end{subfigure}
    \begin{subfigure}{0.32\textwidth}
        \centering
        \includegraphics[width=\textwidth]{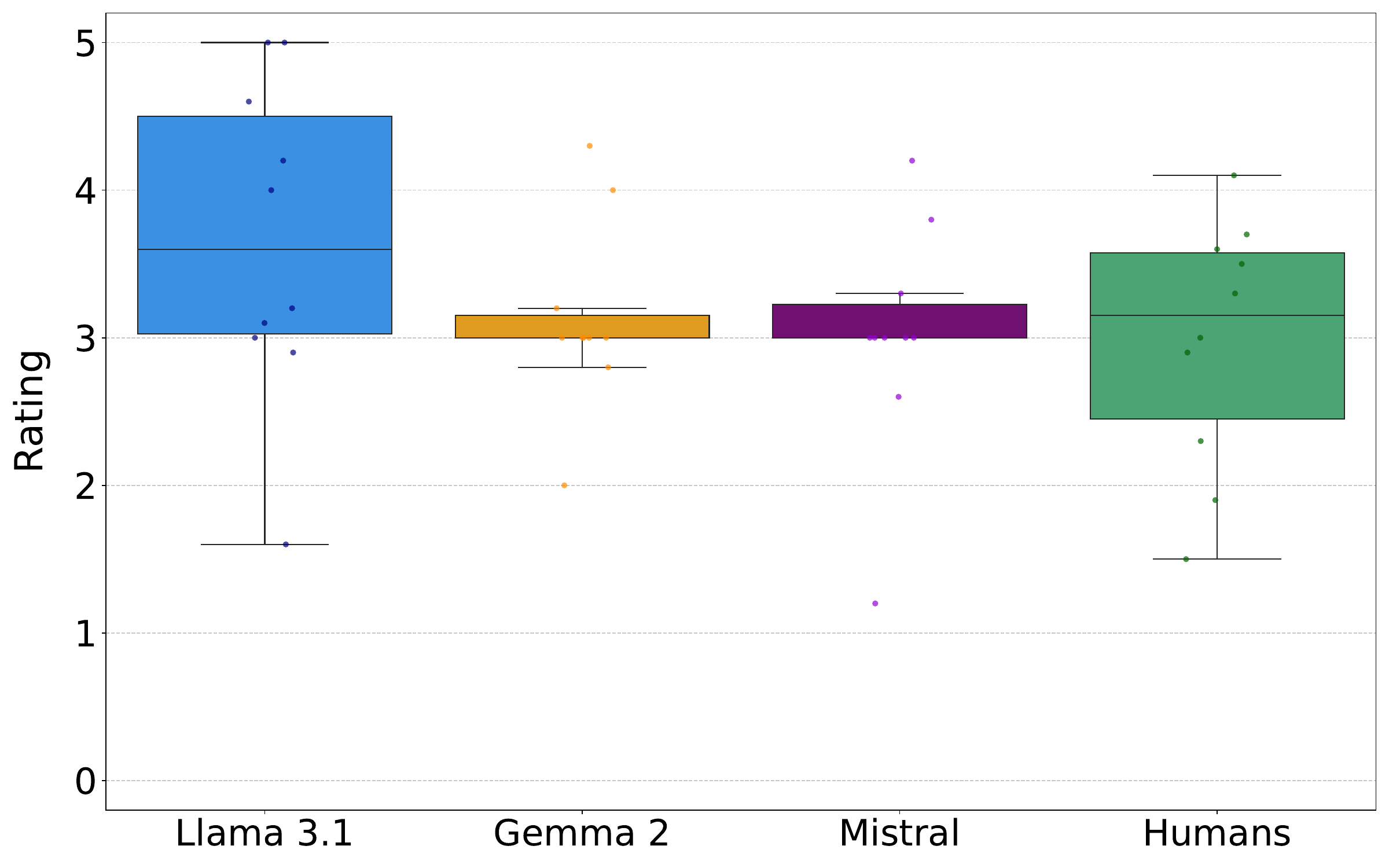}
        \caption{Half-True} \label{fig:Distribution_Ratings_Half_True}
    \end{subfigure}
    \begin{subfigure}{0.32\textwidth}
        \centering
        \includegraphics[width=\textwidth]{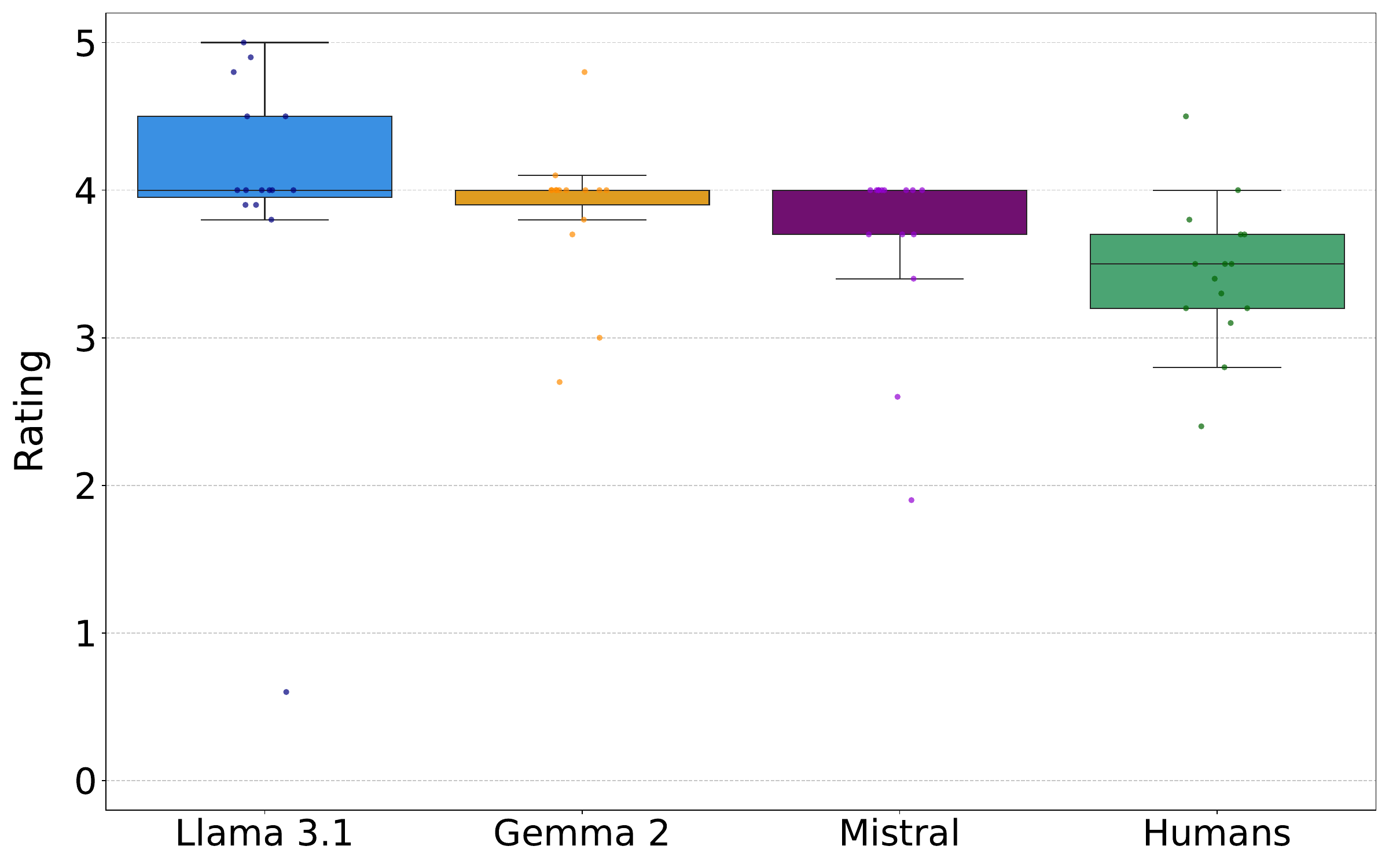}
        \caption{Mostly-True} \label{fig:Distribution_Ratings_Mostly_True}
    \end{subfigure}
    \begin{subfigure}{0.32\textwidth}
        \centering
        \includegraphics[width=\textwidth]{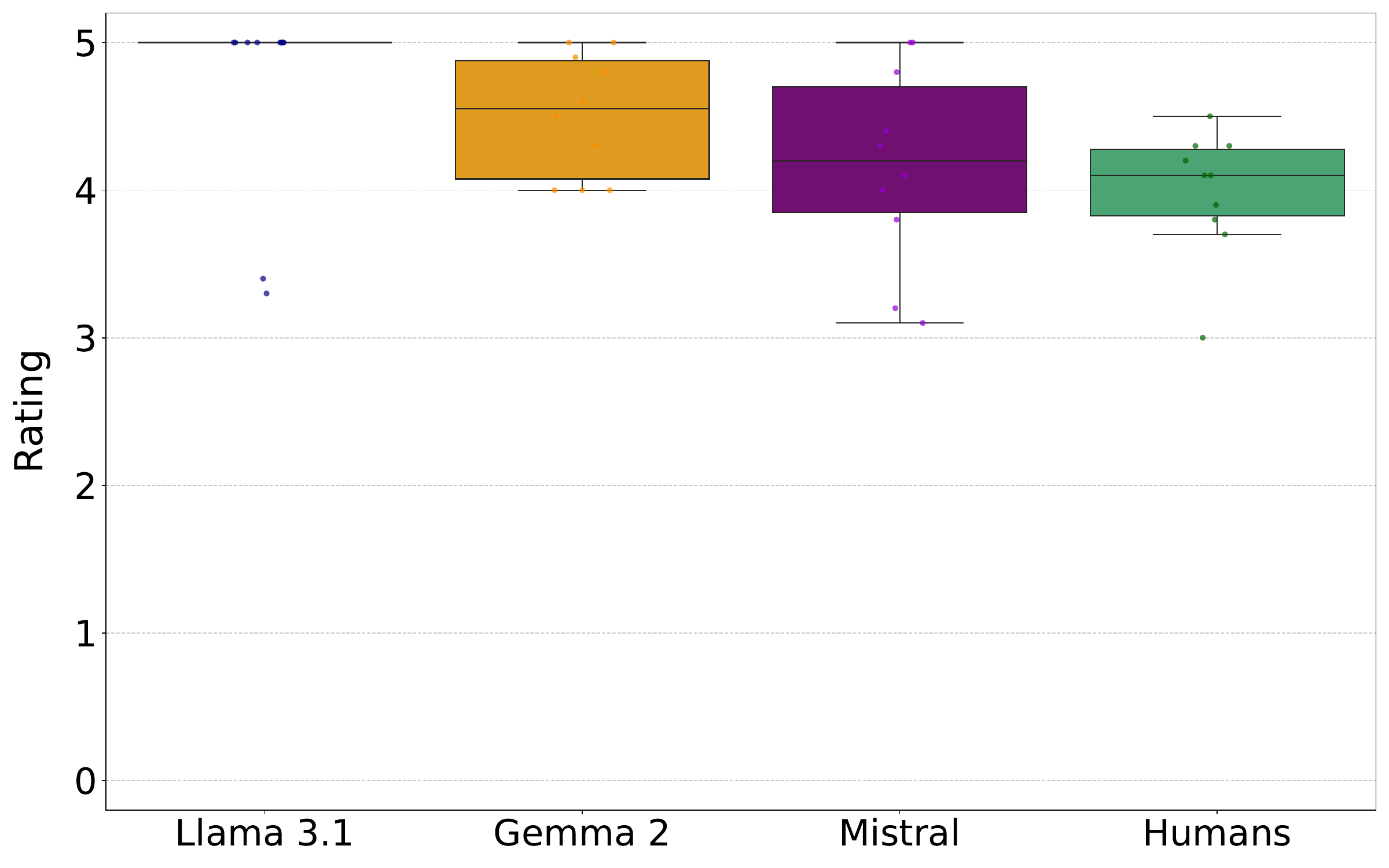}
        \caption{True} \label{fig:Distribution_Ratings_True}
    \end{subfigure}
    \caption{\textbf{Distribution of ratings assigned by generative agents and human workers across different truthfulness levels.} The boxplots highlight the consistency of agent ratings, particularly for extreme labels. In contrast, human workers exhibit wider interquartile ranges and more outliers, indicating higher inconsistency in their assessments.}
    \label{fig:Distribution_Ratings}
\end{figure*}

The distribution of ratings assigned by generative agents and human workers across different evaluation categories provides insights into their assessment consistency and alignment with the ground truth. {\color{black}Figure \ref{fig:Distribution_Ratings} shows the class-wise distributions of ratings, with individual document ratings represented by dots.}

The agents' ratings demonstrate relatively stable alignment with the expected categories, particularly for extreme labels such as \textit{"Pants-on-Fire"} and \textit{"True"}. However, greater variability is observed for intermediate categories such as \textit{"Mostly-False"} for Gemma 2 and Mistral or \textit{"Half-True"} for Llama 3.1, suggesting that these levels are more challenging to assess. By contrast, human workers display significantly wider interquartile ranges and more frequent outliers across most categories, indicating greater inconsistency in their assessments. This is particularly evident in the \textit{"Pants-on-Fire"}, \textit{"Mostly-False"}, \textit{"Mostly-True"}, and \textit{"True"} categories, where a Kruskal-Wallis test revealed statistically significant differences between the human and agent groups ($p-value<0.05$). This contrast highlights the agents' superior stability and precision, particularly for nuanced truthfulness levels.

These findings emphasize the robustness of generative agents in replicating expert-like judgments, with their stability and alignment markedly superior to those of human workers. Moreover, the agents' ability to maintain relatively consistent evaluations, even for challenging intermediate categories, underscores their reliability and potential for more accurate truthfulness assessment.

{\color{black}
\subsubsection{Generative Agents' Fact-Checking Performance by Topic}

To explore whether generative agents exhibit topic-specific variability, we analyzed their performance across three topics: \textit{Civil Rights}, \textit{Conspiracy Theories}, and \textit{Economics}. Table \ref{tab:Performance_By_Topic} shows performance metrics by topic for each generative model across the 2-level and 6-level truthfulness scales.

\begin{table*}[t]
    \centering
    \caption{{\color{black}\textbf{Fact-checking performance of different crowds by claim topic.} Results are reported for both 2-level and 6-level truthfulness classification scales.}}
    \label{tab:Performance_By_Topic}
    {\color{black}
    \resizebox{\linewidth}{!}{%
    \begin{tabular}{lccccccccccc}
        \toprule
        \textbf{Scale} & \textbf{Topic} & \textbf{Crowd} & \textbf{Accuracy} & \textbf{Precision} & \textbf{Recall} & \textbf{F1} & \textbf{Ext. Agreement} & \multicolumn{2}{c}{\textbf{Pairwise Agreement}} & \textbf{Int. Agreement} \\
        \cmidrule(lr){9-10}
        & & & \textbf{(Correct/Total)} & & & & \textbf{\boldmath$\alpha$} & \textbf{Exact} & \textbf{Directional} & \textbf{\boldmath$\alpha$} \\
        \midrule
        \multirow{16}{*}{2-Level}
        & \multirow{4}{*}{Civil Rights}
        & Llama 3.1 & \textbf{0.941 (16/17)} & \textbf{0.949} & \textbf{0.941} & \textbf{0.942} & \textbf{0.881} & - & - & 0.896 \\
        & & Gemma 2 & 0.882 (15/17) & 0.902 & 0.882 & 0.878 & 0.743 & - & - & \textbf{1.000} \\
        & & Mistral & \textbf{0.941 (16/17)} & \textbf{0.949} & \textbf{0.941} & \textbf{0.942} & \textbf{0.881} & - & - & 0.764 \\
        & & Humans & 0.882 (15/17) & 0.882 & 0.882 & 0.882 & 0.758 & - & - & 0.223 \\
        \cmidrule(lr){2-11}
        & \multirow{4}{*}{Conspiracy Theories}
        & Llama 3.1 & \textbf{0.920 (23/25)} & \textbf{0.920} & \textbf{0.920} & \textbf{0.920} & \textbf{0.838} & - & - & 0.879 \\
        & & Gemma 2 & 0.840 (21/25) & 0.883 & 0.840 & 0.839 & 0.680 & - & - & 0.858 \\
        & & Mistral & 0.840 (21/25) & 0.883 & 0.840 & 0.839 & 0.680 & - & - & \textbf{0.890} \\
        & & Humans & 0.880 (22/25) & 0.881 & 0.880 & 0.879 & 0.754 & - & - & 0.134 \\
        \cmidrule(lr){2-11}
        & \multirow{4}{*}{Economics}
        & Llama 3.1 & \textbf{1.000 (28/28)} & \textbf{1.000} & \textbf{1.000} & \textbf{1.000} & \textbf{1.000} & - & - & 0.874 \\
        & & Gemma 2 & 0.929 (26/28) & \textbf{0.929} & 0.929 & 0.929 & 0.857 & - & - & \textbf{0.962} \\
        & & Mistral & 0.929 (26/28) & 0.929 & 0.929 & 0.929 & 0.857 & - & - & 0.856 \\
        & & Humans & 0.893 (25/28) & 0.895 & 0.893 & 0.893 & 0.786 & - & - & 0.168 \\
        \cmidrule{1-11}
        \cmidrule{1-11}
        \cmidrule{1-11}
        \multirow{12}{*}{6-Level}
        & \multirow{4}{*}{Civil Rights}
        & Llama 3.1 & \textbf{0.647 (11/17)} & \textbf{0.556} & \textbf{0.647} & \textbf{0.580} & 0.888 & 0.026 & \textbf{0.231} & 0.857 \\
        & & Gemma 2 & 0.471 (8/17) & 0.424 & 0.471 & 0.429 & \textbf{0.900} & \textbf{0.179} & \textbf{0.231} & \textbf{0.951} \\
        & & Mistral & 0.353 (6/17) & 0.515 & 0.353 & 0.412 & 0.750 & 0.026 & \textbf{0.231} & 0.800 \\
        & & Humans & 0.353 (6/17) & 0.315 & 0.353 & 0.327 & 0.694 & 0.026 & 0.179 & 0.358 \\
        \cmidrule(lr){2-11}
        & \multirow{4}{*}{Conspiracy Theories}
        & Llama 3.1 & \textbf{0.680 (17/25)} & \textbf{0.660} & \textbf{0.680} & \textbf{0.624} & 0.743 & \textbf{0.218} & \textbf{0.346} & 0.899 \\
        & & Gemma 2 & 0.440 (11/25) & 0.531 & 0.440 & 0.385 & 0.691 & 0.090 & 0.269 & \textbf{0.926} \\
        & & Mistral & 0.440 (11/25) & 0.373 & 0.440 & 0.341 & 0.653 & 0.090 & 0.256 & 0.912 \\
        & & Humans & 0.560 (14/25) & 0.598 & 0.560 & 0.534 & \textbf{0.845} & 0.128 & 0.269 & 0.227 \\
        \cmidrule(lr){2-11}
        & \multirow{4}{*}{Economics}
        & Llama 3.1 & 0.464 (13/28) & \textbf{0.602} & 0.464 & \textbf{0.458} & 0.727 & 0.058 & 0.389 & 0.838 \\
        & & Gemma 2 & \textbf{0.500 (14/28)} & 0.500 & \textbf{0.500} & 0.451 & 0.898 & \textbf{0.231} & \textbf{0.433} & \textbf{0.942} \\
        & & Mistral & 0.464 (13/28) & 0.402 & 0.464 & 0.385 & 0.813 & 0.092 & 0.292 & 0.852 \\
        & & Humans & 0.464 (13/28) & 0.481 & \textbf{0.464} & 0.454 & 0.791 & 0.197 & 0.397 & 0.261 \\
        \bottomrule
    \end{tabular}%
    }
    }
\end{table*}

The results reveal several noteworthy trends. Regardless of the annotation scheme and the topic, crowds of generative agents perform better than human crowd in terms of accuracy, precision, recall and F1-score. In particular, agents powered by Llama 3.1 confirm to be the best performer. For example, on the 6-level scale, they exhibit F1-score improvements over the second best configuration of 35.2\%, 16.4\%, and 1.5\% for \textit{"Civil Rights"}, \textit{"Conspiracy Theories"}, and \textit{"Economics"}, respectively. 

In addition, while topic-wise performance are stable on the 2-level scale, we observe larger differences on the 6-level scale. For instance, Llama 3.1 achieves accuracy equals to 0.647 on claims related to \textit{"Civil Rights"} but only 0.464 on \textit{"Economics"}, which suggests that fine-grained veracity distinctions are more challenging in domains involving technical or quantitative content.

Finally, in terms of agreement, we observe that claims on \textit{"Civil Rights"} yield the highest external agreement scores for both Llama 3.1 ($\alpha = 0.888$) and Gemma 2 ($\alpha = 0.900$), while pairwise agreement is generally weaker for \textit{"Conspiracy Theories"}, possibly reflecting the inherently ambiguous or polarized nature of such claims. Internal consistency remains high across all topics, confirming that generative agents tend to apply evaluation criteria in a stable and coherent manner, even when accuracy varies.

Overall, this analysis underscores the importance of considering topical context when deploying generative agents in real-world fact-checking scenarios. While agents show strong general performance-consistently matching or exceeding the performance of human annotators-their effectiveness may vary depending on the content domain, particularly when fine-grained distinctions are required.
}

{\color{black}
\subsubsection{Effect of Varying the Number of Agents per Statement}

\begin{table*}[t]
    \centering
    \caption{{\color{black}\textbf{Fact-checking performance varying the number of generative agents evaluating each statement.} Generative agents maintain high accuracy and agreement even when operating in smaller groups (3 agents), especially on binary (2-level) tasks. On more complex 6-level classifications, performance improves with larger crowds, but plateaus beyond 10 agents, indicating that additional redundancy yields minimal gains.}}
    \label{tab:Varying_Agents_Results}
    \resizebox{\linewidth}{!}{%
    {\color{black}
    \begin{tabular}{lcccccccccc}
        \toprule
        \textbf{Scale} & \textbf{Crowd} & \textbf{\#Agents} & \textbf{Accuracy} & \textbf{Precision} & \textbf{Recall} & \textbf{F1} & \textbf{Ext. Agreement \boldmath$\alpha$} & \multicolumn{2}{c}{\textbf{Pairwise Agreement}} & \textbf{Int. Agreement} \\
        \cmidrule(lr){9-10}
        & & & \textbf{(Correct/Total)} & & & & \textbf{\boldmath$\alpha$} & \textbf{Exact} & \textbf{Directional} & \textbf{\boldmath$\alpha$}\\
        \midrule
        \multirow{9}{*}{2-Level}
        & \multirow{3}{*}{Llama 3.1}
        & 3  & 0.929 (65/70) & 0.968 & 0.885 & 0.925 & 0.858 & -- & -- & 0.810 \\
        & & 10 & \textbf{0.957 (67/70)} & \textbf{0.970} & \textbf{0.942} & \textbf{0.956} & \textbf{0.914} & -- & -- & \textbf{0.881} \\
        & & 15 & \textbf{0.957 (67/70)} & \textbf{0.970} & \textbf{0.942} & \textbf{0.956} & \textbf{0.914} & - & - & \textbf{0.881} \\
        \cmidrule(lr){2-11}
        & \multirow{3}{*}{Gemma 2}
        & 3  & 0.900 (63/70) & \textbf{0.850} & 0.971 & 0.906 & \textbf{0.800} & -- & -- & 0.883 \\
        & & 10 & 0.885 (62/70) & 0.829 & 0.971 & 0.895 & 0.771 & -- & -- & 0.933 \\
        & & 15 & \textbf{0.914 (64/70)} & 0.832 & \textbf{0.979} & \textbf{0.925} & 0.791 & - & - & \textbf{0.939} \\
        \cmidrule(lr){2-11}
        & \multirow{3}{*}{Mistral}
        & 3  & \textbf{0.914 (64/70)} & \textbf{0.891} & \textbf{0.943} & \textbf{0.917} & \textbf{0.823} & -- & -- & 0.787 \\
        & & 10 & 0.900 (63/70) & 0.868 & \textbf{0.943} & 0.904 & 0.801 & -- & -- & \textbf{0.845} \\
        & & 15 & 0.900 (63/70) & 0.868 & \textbf{0.943} & 0.904 & 0.801 & - & - & \textbf{0.845} \\
        \cmidrule{1-11}
        \cmidrule{1-11}
        \cmidrule{1-11}
        \multirow{9}{*}{6-Level}
        & \multirow{3}{*}{Llama 3.1}
        & 3  & 0.543 (38/70) & 0.613 & 0.540 & 0.530 & \textbf{0.770} & \textbf{0.355} & 0.707 & 0.777 \\
        & & 10 & 0.585 (41/70) & 0.619 & 0.585 & 0.548 & 0.759 & 0.331 & 0.714 & 0.867 \\
        & & 15 & \textbf{0.600 (42/70)} & \textbf{0.623} & \textbf{0.591} & \textbf{0.553} & 0.762 & 0.336 & \textbf{0.718} & \textbf{0.872} \\
        \cmidrule(lr){2-11}
        & \multirow{3}{*}{Gemma 2}
        & 3  & 0.457 (32/70) & 0.467 & 0.486 & \textbf{0.433} & 0.803 & 0.331 & \textbf{0.750} & 0.907 \\
        & & 10 & \textbf{0.471 (33/70)} & \textbf{0.626} & \textbf{0.471} & \textbf{0.433} & \textbf{0.820} & \textbf{0.335} & 0.748 & 0.945 \\
        & & 15 & \textbf{0.471 (33/70)} & \textbf{0.626} & \textbf{0.471} & \textbf{0.433} & \textbf{0.820} & \textbf{0.335} & 0.748 & \textbf{0.955} \\
        \cmidrule(lr){2-11}
        & \multirow{3}{*}{Mistral}
        & 3  & \textbf{0.457 (32/70)} & \textbf{0.588} & \textbf{0.481} & \textbf{0.425} & \textbf{0.810} & \textbf{0.330} & \textbf{0.741} & 0.891 \\
        & & 10 & 0.428 (30/70) & 0.399 & 0.428 & 0.370 & 0.745 & 0.285 & 0.690 & \textbf{0.907} \\
        & & 15 & 0.428 (30/70) & 0.399 & 0.428 & 0.370 & 0.735 & 0.285 & 0.690 & \textbf{0.907} \\
        \bottomrule
    \end{tabular}
    }
    }
\end{table*}

We investigate how performance varies with respect to the number of generative agents in the crowd. Specifically, in the main experimental setup, each claim was independently reviewed by 10 generative agents. Here, we examine two alternative scenarios: a reduced configuration in which only 3 agents assess each statement, and an expanded one where the crowd annotating each claim is composed by 15 agents. Table \ref{tab:Varying_Agents_Results} reports the results obtained under all settings.

With the exception of generative agents powered by Mistral, we observe that the 3-agent setting always yields worse performance than the original 10-agent configuration. This gap is especially pronounced in the 6-level classification scheme. For example, crowds powered by Llama 3.1 on the 6-level scale achieves an accuracy of 0.543, which represents a 6.8\% decline over the 0.585 achieved under the 10-agent setting. Nevertheless, internal and external agreement remained strong, indicating that generative agents produce coherent judgments even with limited aggregation. This result highlights the need for larger crowds to reach consensus on fine-grained fact-checking classification.

Furthermore, we observe identical or marginal improvements of the 15-agent configuration over the 10-agent setup. For example, agents powered by Gemma 2 achieved the same performance on the 6-level scale, and for those powered by Llama 3.1, accuracy increased from 0.585 with 10 agents to 0.600 with 15 agents—corresponding to a relative increase of approximately 2.5\%. These results suggest that, beyond a certain threshold, additional agents do not significantly enhance classification performance.

Overall, these findings reveal a key pattern: generative agent crowds exhibit aggregation dynamics analogous to human-based crowdsourcing, where few evaluators are sufficient for simpler binary (2-level) classifications, whereas more complex (6-level) veracity judgments benefit from larger crowds, up to a plateau beyond which performance stabilizes and further agents offer negligible gains.
}

\begin{tcolorbox}[colback=gray!10, colframe=gray!80, title=Takeaway (RQ1)]
{\color{black} Generative agents consistently outperform human annotators in both accuracy and agreement for crowdsourced fact-checking.

}
\end{tcolorbox}

{\color{black}
\subsection{RQ2: What is the role of external evidence and claim recency in generative agents' fact-checking performance?}

To better understand the capabilities of generative agents in crowdsourced fact-checking, we investigate how their performance is affected by two key factors: \textit{(i)} the influence of the news selected as evidence to verify the claim, and \textit{(ii)} the influence of claim recency.

}

{\color{black}
\subsubsection{Effect of the External Evidence}

To isolate the contribution of the \textit{Evidence Selection} phase in our fact-checking pipeline, we conducted an ablation experiment in which generative agents were tasked with evaluating the truthfulness of each claim using only their internal knowledge, without access to any external evidence. 

As shown in Table \ref{tab:No_Evidence_Results}, the accuracy of generative agents declines markedly in the absence of external evidence compared to the evidence-informed setting that we have considered in previous analyses (see Table \ref{tab:RQ1_Results}). Specifically, on the 2-level scale, Llama 3.1 drops by -26.9\%, Mistral by -23.8\%, and Gemma 2 by -16.0\%. The effect is even more pronounced on the 6-level scale: Gemma 2 and Llama 3.1 experience -45.4\% and -48.7\% reduction in accuracy, respectively, while Mistral suffers a drop of -63.3\%. We observe the same trend for external agreement metrics: for example, on the 2-level scale, Llama 3.1 drops from $\alpha = 0.914$ (with evidence) to $\alpha = 0.403$ (without), corresponding to a 55.9\% reduction. Gemma 2 decreases from $\alpha = 0.771$ to $\alpha = 0.489$ (-36.6\%), and Mistral from $\alpha = 0.801$ to $\alpha = 0.363$ (-54.7\%).

\begin{table*}[t]
    \centering
    \caption{{\color{black}\textbf{Fact-checking performance of generative agents without evidence.} Results show that, while agents maintain moderate performance on the 2-level scale, their effectiveness significantly deteriorates on the more granular 6-level scale.}}
    \label{tab:No_Evidence_Results}
    \resizebox{\linewidth}{!}{%
    {\color{black}
    \begin{tabular}{lccccccccccc}
        \toprule
        \textbf{Scale} & \textbf{Crowd} & \textbf{Accuracy} & \textbf{Precision} & \textbf{Recall} & \textbf{F1} & \textbf{Ext. Agreement} & \multicolumn{2}{c}{\textbf{Pairwise Agreement}} & \textbf{Int. Agreement} \\
        \cmidrule(lr){8-9}
        & & \textbf{(Correct/Total)} & & & & \textbf{\boldmath$\alpha$} & \textbf{Exact} & \textbf{Directional} & \textbf{\boldmath$\alpha$} \\
        \midrule
        2-Level & \begin{tabular}{@{}c@{}}Llama 3.1 \\ Gemma 2 \\ Mistral \end{tabular} & 
        \begin{tabular}{@{}c@{}} 0.700 (49/70) \\ \textbf{0.743 (52/70)} \\ 0.686 (48/70) \end{tabular} & 
        \begin{tabular}{@{}c@{}} 0.719 \\ 0.730 \\ \textbf{0.760} \end{tabular} &
        \begin{tabular}{@{}c@{}} 0.657 \\ \textbf{0.771} \\ 0.543 \end{tabular} &
        \begin{tabular}{@{}c@{}} 0.686 \\ \textbf{0.750} \\ 0.633 \end{tabular} &
        \begin{tabular}{@{}c@{}} 0.403 \\ \textbf{0.489} \\ 0.363 \end{tabular} &
        \begin{tabular}{@{}c@{}}-- \\ -- \\ -- \end{tabular} &
        \begin{tabular}{@{}c@{}}-- \\ -- \\ -- \end{tabular} &
        \begin{tabular}{@{}c@{}} 0.812 \\ \textbf{0.816} \\ 0.720 \end{tabular} \\
        \midrule
        6-Level & \begin{tabular}{@{}c@{}}Llama 3.1 \\ Gemma 2 \\ Mistral \end{tabular} & 
        \begin{tabular}{@{}c@{}} \textbf{0.300 (21/70)} \\ 0.257 (18/70) \\ 0.157 (11/70) \end{tabular} & 
        \begin{tabular}{@{}c@{}} 0.254 \\ \textbf{0.359} \\ 0.315 \end{tabular} &
        \begin{tabular}{@{}c@{}} \textbf{0.283} \\ 0.281 \\ 0.188 \end{tabular} &
        \begin{tabular}{@{}c@{}} \textbf{0.265} \\ 0.244 \\ 0.135 \end{tabular} &
        \begin{tabular}{@{}c@{}} 0.446 \\ \textbf{0.535} \\ 0.364 \end{tabular} &
        \begin{tabular}{@{}c@{}} 0.167 \\ \textbf{0.200} \\ 0.181 \end{tabular} &
        \begin{tabular}{@{}c@{}} 0.539 \\ \textbf{0.590} \\ 0.508 \end{tabular} &
        \begin{tabular}{@{}c@{}} 0.806 \\ \textbf{0.875} \\ 0.824 \end{tabular} \\
        \bottomrule
    \end{tabular}
    }
    }
\end{table*}

These findings highlight the critical role of external evidence in enabling generative agents to perform accurate and consistent fact-checking, particularly when faced with fine-grained or ambiguous claims. This aligns with recent work on retrieval-augmented generation (RAG), which shows that conditioning LLMs on retrieved or provided documents improves their reliability, factuality, and coherence \cite{chen2024benchmarking, asai2023self}.
}

{\color{black}
\subsubsection{Role of Claim Recency}

To further assess the generalizability of our approach, we conducted a series of experiments to assess how the claim recency influences the performance of generative agents in crowdsourced fact-checking. 

\paragraph*{Older Claims}
We compare performance achieved by generative agents on the 70 claims published in 2022, with another set of 70 claims in the original dataset \cite{Crowd_Fact_Checking_Base_Paper} published between 2008 and 2017. This evaluation aimed to assess whether temporally outdated content introduces systematic challenges regardless of the evidence selected for the evaluation process. The results, reported in Table \ref{tab:Old_Claims_Results}, indicate a marked decline in agent performance on older claims for all generative agents—particularly on the 6-level truthfulness scale. For instance, Llama 3.1, which previously achieved an F1 of 0.548 and an accuracy of 0.585 on 2022 claims (see Table \ref{tab:RQ1_Results}), drops to an F1 of 0.247 and an accuracy of 0.257 on this older subset. Interestingly, human crowd suffers from a similar performance drop: the F1 score decreases from 0.471 to 0.214 on the 6-level scale. This suggests that older claims are more likely to suffer from contextual drift (e.g., policy changes, evolving terminology), and the factual relevance of supporting evidence may no longer be verifiable in present-day discourse.

\begin{table*}[t]
    \centering
    \caption{{\color{black}\textbf{Performance of generative agents on older claims (2008–2017).} Results show a marked drop in performance compared to the results obtained on 2022 claims, particularly on the 6-level scale.}}
    \label{tab:Old_Claims_Results}
    {\color{black}
    \resizebox{\linewidth}{!}{%
    \begin{tabular}{lccccccccccc}
        \toprule
        \textbf{Scale} & \textbf{Crowd} & \textbf{Accuracy} & \textbf{Precision} & \textbf{Recall} & \textbf{F1} & \textbf{Ext. Agreement} & \multicolumn{2}{c}{\textbf{Pairwise Agreement}} & \textbf{Int. Agreement} \\
        \cmidrule(lr){8-9}
        & & \textbf{(Correct/Total)} & & & & \textbf{\boldmath$\alpha$} & \textbf{Exact} & \textbf{Directional} & \textbf{\boldmath$\alpha$} \\
        \midrule
        2-Level & \begin{tabular}{@{}c@{}}Llama 3.1 \\ Gemma 2 \\ Mistral \\ Humans\end{tabular} & 
        \begin{tabular}{@{}c@{}}0.657 (46/70) \\ \textbf{0.771 (54/70)} \\ 0.729 (51/70) \\ 0.471 (33/70)\end{tabular} & 
        \begin{tabular}{@{}c@{}}0.684 \\ \textbf{0.750} \\ 0.711 \\ 0.677\end{tabular}  &
        \begin{tabular}{@{}c@{}}0.684 \\ 0.868 \\ 0.842 \\ \textbf{0.913}\end{tabular} &
        \begin{tabular}{@{}c@{}}0.684 \\ \textbf{0.804} \\ 0.771 \\ 0.778 \end{tabular} &
        \begin{tabular}{@{}c@{}}0.314 \\ \textbf{0.532} \\ 0.441 \\ 0.451\end{tabular} &
        \begin{tabular}{@{}c@{}}-- \\ -- \\ -- \\ --\end{tabular} &
        \begin{tabular}{@{}c@{}}-- \\ -- \\ -- \\ --\end{tabular} &
        \begin{tabular}{@{}c@{}}0.658 \\ \textbf{0.825} \\ 0.803 \\ 0.125\end{tabular} \\
        \midrule
        6-Level & \begin{tabular}{@{}c@{}}Llama 3.1 \\ Gemma 2 \\ Mistral \\ Humans\end{tabular} & 
        \begin{tabular}{@{}c@{}}0.257 (18/70) \\ \textbf{0.400 (28/70)} \\ 0.357 (25/70) \\ 0.214 (15/70)\end{tabular} & 
        \begin{tabular}{@{}c@{}}0.262 \\ \textbf{0.531} \\ 0.434 \\ 0.376\end{tabular} &
        \begin{tabular}{@{}c@{}}0.252 \\ \textbf{0.382} \\ 0.339 \\ 0.346\end{tabular} &
        \begin{tabular}{@{}c@{}}0.247 \\ \textbf{0.380} \\ 0.286 \\ 0.343\end{tabular} &
        \begin{tabular}{@{}c@{}}0.445 \\ \textbf{0.629} \\ 0.503 \\ 0.292\end{tabular} &
        \begin{tabular}{@{}c@{}}0.172 \\ \textbf{0.256} \\ 0.237 \\ 0.107\end{tabular} &
        \begin{tabular}{@{}c@{}}0.538 \\ \textbf{0.563} \\ 0.524 \\ 0.373\end{tabular} &
        \begin{tabular}{@{}c@{}}0.804 \\ \textbf{0.926} \\ 0.856 \\ 0.180\end{tabular} \\
        \bottomrule
    \end{tabular}
    }
    }
\end{table*}

\paragraph*{Newest Claims}

We conducted a second experiment using a different set of 70 claims extracted from \cite{Factors_Dataset}, all published in 2025\footnote{These claims cover the same topics-civil rights, economics, conspiracy theories-identified in the original dataset \cite{Crowd_Fact_Checking_Base_Paper}. An example of claim released on January 2025: \textit{“The Biden administration is offering only \$770 for people who lost everything in the Los Angeles wildfires”. }}. These claims were selected to ensure they post-date the training cut-off of all LLMs employed in our experiments, ensuring that agents could not rely on memorized knowledge and were required to base their veracity assessments entirely on evidence. This setting provides a stronger guarantee that performance reflects the agents’ ability to interpret and reason over the provided evidence rather than memorize facts seen during training, avoiding the possibility of data leakage, i.e., agents having prior exposure to the claims and their veracity labels \cite{carlini2021extracting}.


Table \ref{tab:NoLeakage_Results} shows fact-checking performance for crowds powered Llama 3.1, Gemma 2 and Mistral. Notably, we do not report performance of the human crowd as these claims were not part of the original study on crowdsourced fact-checking \cite{Crowd_Fact_Checking_Base_Paper}. We observe that generative agents maintained strong performance even on this set of very recent claims. Similar to previous analyses, Llama 3.1 achieved the highest accuracy (0.943) on the 2-level scale, followed closely by Mistral (0.900) and Gemma 2 (0.886). Similar trends held for the more challenging 6-level classification, with F1 scores and agreement metrics comparable to those obtained on the 2022 benchmark set (see Table \ref{tab:RQ1_Results}). These results demonstrate that when supported by relevant and up-to-date evidence, generative agents can reliably verify claims that fall entirely outside their training window.


\begin{table*}[t]
    \centering
    \caption{{\color{black}\textbf{Performance metrics for fact-checking on claims published on 2025, after the training cut-off dates of all LLMs employed.} Generative agents maintain strong performance when verifying novel claims unavailable during model training, confirming the effectiveness of evidence-guided reasoning.}}
    \label{tab:NoLeakage_Results}
    \resizebox{\linewidth}{!}{%
    {\color{black}
    \begin{tabular}{lccccccccccc}
        \toprule
        \textbf{Scale} & \textbf{Crowd} & \textbf{Accuracy} & \textbf{Precision} & \textbf{Recall} & \textbf{F1} & \textbf{Ext. Agreement} & \multicolumn{2}{c}{\textbf{Pairwise Agreement}} & \textbf{Int. Agreement} \\
        \cmidrule(lr){8-9}
        & & \textbf{(Correct/Total)} & & & & \textbf{\boldmath$\alpha$} & \textbf{Exact} & \textbf{Directional} & \textbf{\boldmath$\alpha$}\\
        \midrule
        2-Level & \begin{tabular}{@{}c@{}}Llama 3.1 \\ Gemma 2 \\ Mistral\end{tabular} & 
        \begin{tabular}{@{}c@{}}\textbf{0.943 (66/70)} \\ 0.886 (62/70) \\ 0.900 (63/70)\end{tabular} & 
        \begin{tabular}{@{}c@{}}\textbf{0.961} \\ 0.819 \\ 0.871\end{tabular} &
        \begin{tabular}{@{}c@{}}0.914 \\ \textbf{0.943} \\ 0.940\end{tabular} &
        \begin{tabular}{@{}c@{}}\textbf{0.937} \\ 0.877 \\ 0.905\end{tabular} &
        \begin{tabular}{@{}c@{}}\textbf{0.901} \\ 0.754 \\ 0.783\end{tabular} &
        \begin{tabular}{@{}c@{}}-- \\ -- \\ --\end{tabular} &
        \begin{tabular}{@{}c@{}}-- \\ -- \\ --\end{tabular} &
        \begin{tabular}{@{}c@{}}0.877 \\ \textbf{0.922} \\ 0.832\end{tabular} \\
        \midrule
        6-Level & \begin{tabular}{@{}c@{}}Llama 3.1 \\ Gemma 2 \\ Mistral\end{tabular} & 
        \begin{tabular}{@{}c@{}}\textbf{0.543 (38/70)} \\ 0.429 (30/70) \\ 0.457 (32/70)\end{tabular} & 
        \begin{tabular}{@{}c@{}}0.598 \\ \textbf{0.612} \\ 0.421\end{tabular} &
        \begin{tabular}{@{}c@{}}\textbf{0.543} \\ 0.429 \\ 0.457\end{tabular} &
        \begin{tabular}{@{}c@{}}\textbf{0.523} \\ 0.416 \\ 0.386\end{tabular} &
        \begin{tabular}{@{}c@{}}0.741 \\ \textbf{0.788} \\ 0.722\end{tabular} &
        \begin{tabular}{@{}c@{}}0.321 \\ \textbf{0.328} \\ 0.293\end{tabular} &
        \begin{tabular}{@{}c@{}}0.698 \\ \textbf{0.731} \\ 0.679\end{tabular} &
        \begin{tabular}{@{}c@{}}0.854 \\ \textbf{0.934} \\ 0.895\end{tabular} \\
        \bottomrule
    \end{tabular}
    }
    }
\end{table*}
}

\begin{tcolorbox}[colback=gray!10, colframe=gray!80, title=Takeaway (RQ2)]
Generative agents successfully leverage external evidence to assess claim truthfulness. 
\end{tcolorbox}

{\color{black}
\subsection{RQ3: Do humans and generative agents rely on the same factors when assessing truthfulness?}

Here, we examine the factors that influence truthfulness assessments and compare how these factors affect human and generative agent judgments. We differentiate between claim-related characteristics and user-specific attributes. For the former, we analyze how humans and generative agents evaluated the seven quality dimensions described in Section \ref{sec:simulation_workflow} prior to making their final truthfulness decisions. For user-specific attributes, we assess the accuracy of both groups across different demographic and ideological profiles.
}


\subsubsection{Claim-Related Characteristics}

To comprehensively investigate the factors most correlated with truthfulness for generative agents and humans, generative agents were prompted to evaluate the set of statements across various quality dimensions. Human evaluations for the same statements were extracted from the dataset of \cite{Crowd_Fact_Checking_Base_Paper}.

For each quality dimension and for both groups (agents and humans), we computed the frequency of ratings and the correlation between truthfulness and the selected quality dimension. This analysis adopts an exploratory perspective and does not assume a conceptual or causal link between the such dimensions and truthfulness. Rather, it aims to assess whether these commonly used quality indicators are implicitly employed—by either humans or agents—as cues in truthfulness judgments. The objective is not to claim that attributes such as completeness or precision define veracity, but to examine whether their presence systematically co-occurs with higher truthfulness ratings. Additionally, correlation serves to compare evaluative strategies across groups, offering insight into the degree of alignment or divergence between human and agent reasoning patterns.

\paragraph*{Frequency}

\begin{figure*}[htbp]
    \centering
    \begin{subfigure}{0.32\textwidth}
        \centering
        \includegraphics[width=\textwidth]{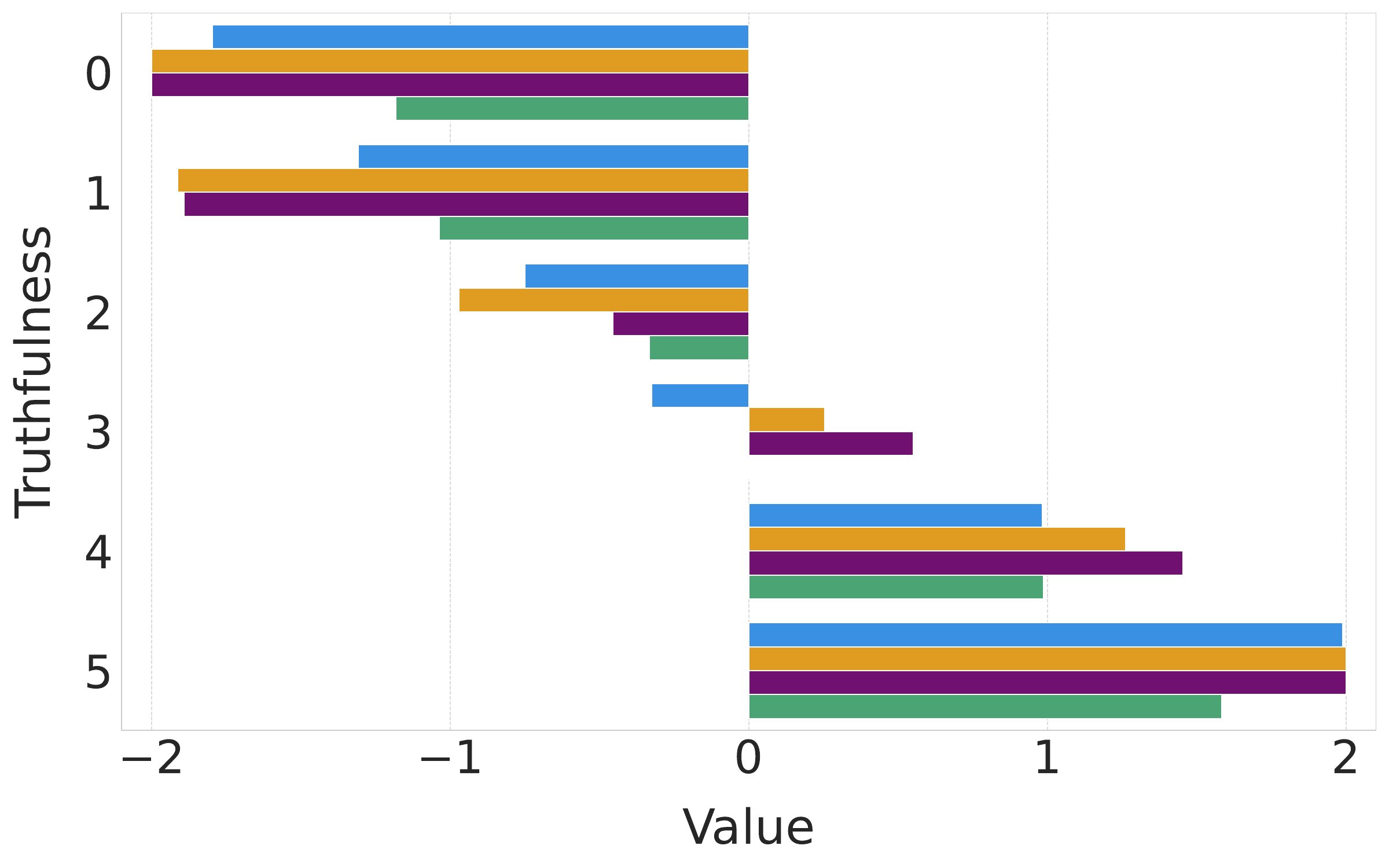}
        \caption{Accuracy} \label{fig:Dimensions_Accuracy}
    \end{subfigure}
    \begin{subfigure}{0.32\textwidth}
        \centering
        \includegraphics[width=\textwidth]{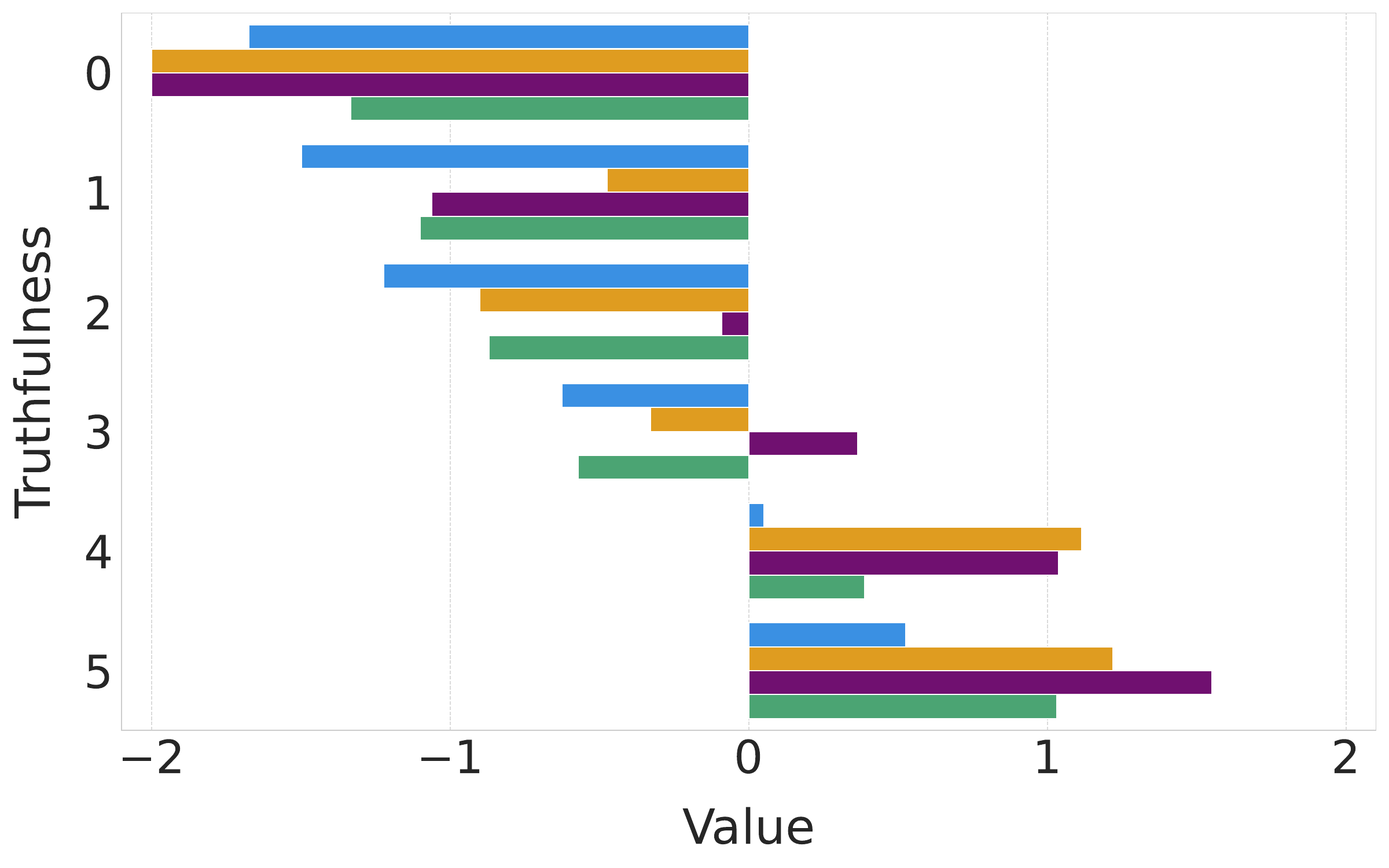}
        \caption{Unbiasedness} \label{fig:Dimensions_Unbiasedness}
    \end{subfigure}
    \begin{subfigure}{0.32\textwidth}
        \centering
        \includegraphics[width=\textwidth]{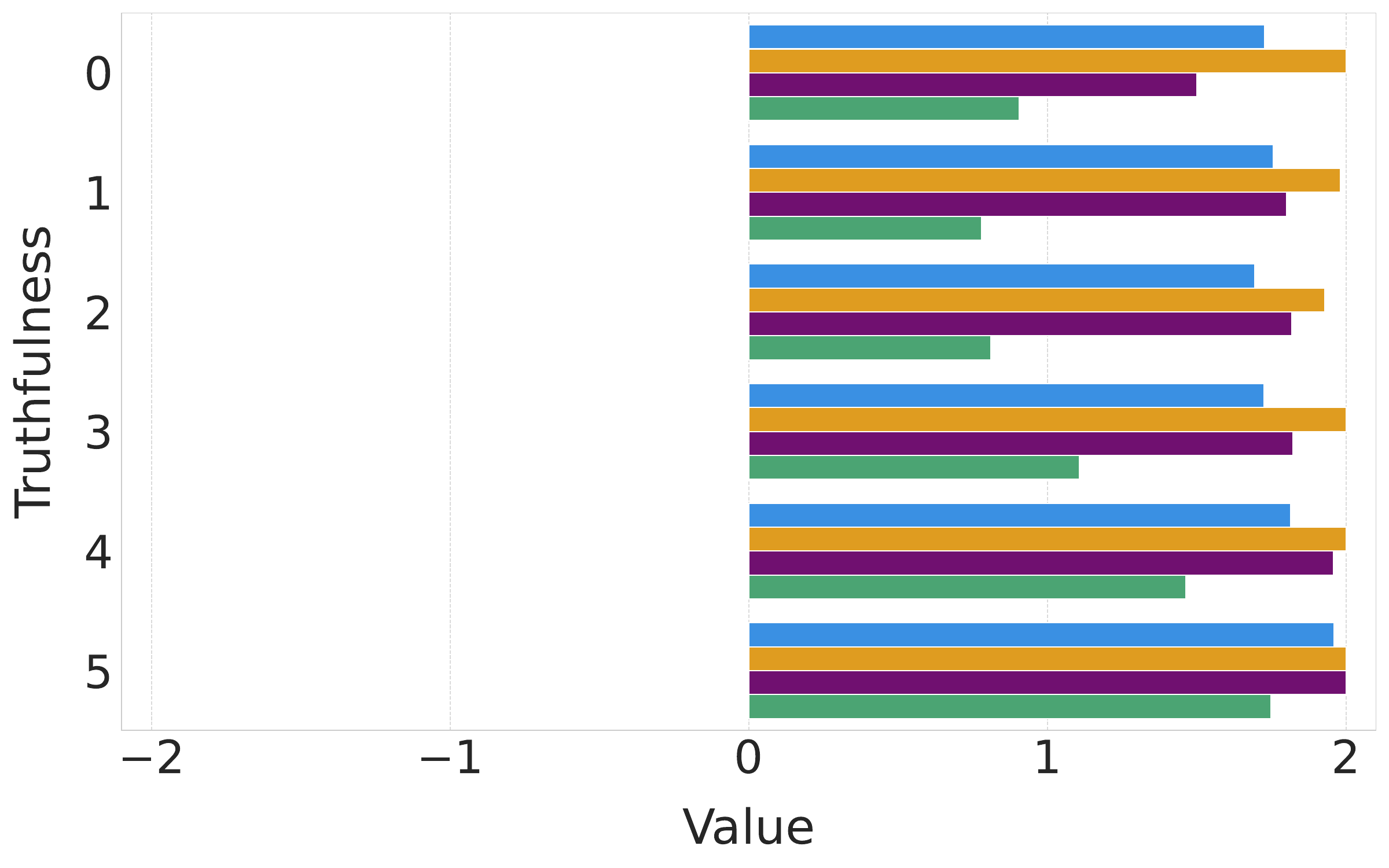}
        \caption{Comprehensibility} \label{fig:Dimensions_Comprehensibility}
    \end{subfigure}
    
    \begin{subfigure}{0.32\textwidth}
        \centering
        \includegraphics[width=\textwidth]{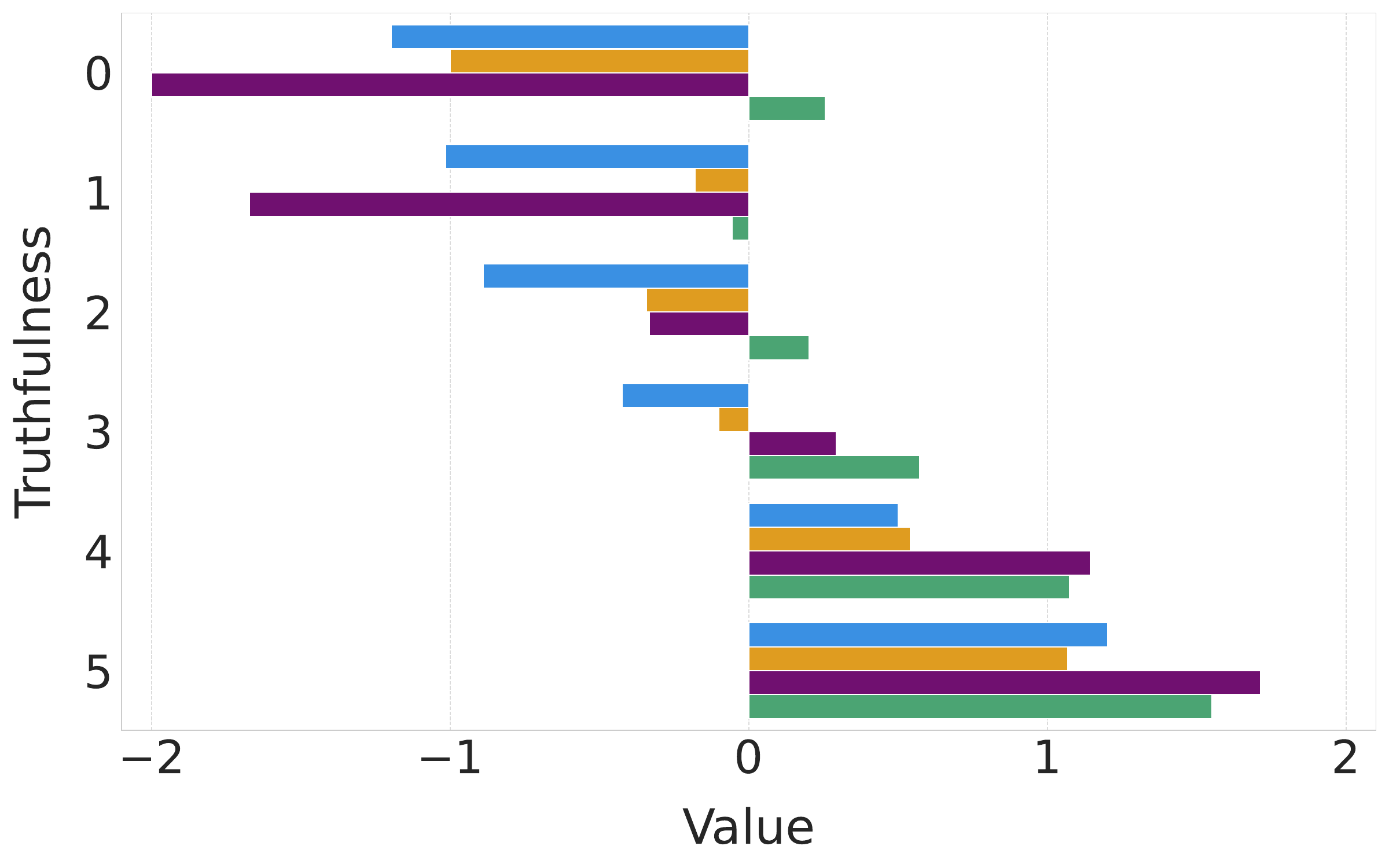}
        \caption{Precision} \label{fig:Dimensions_Precision}
    \end{subfigure}
    \begin{subfigure}{0.32\textwidth}
        \centering
        \includegraphics[width=\textwidth]{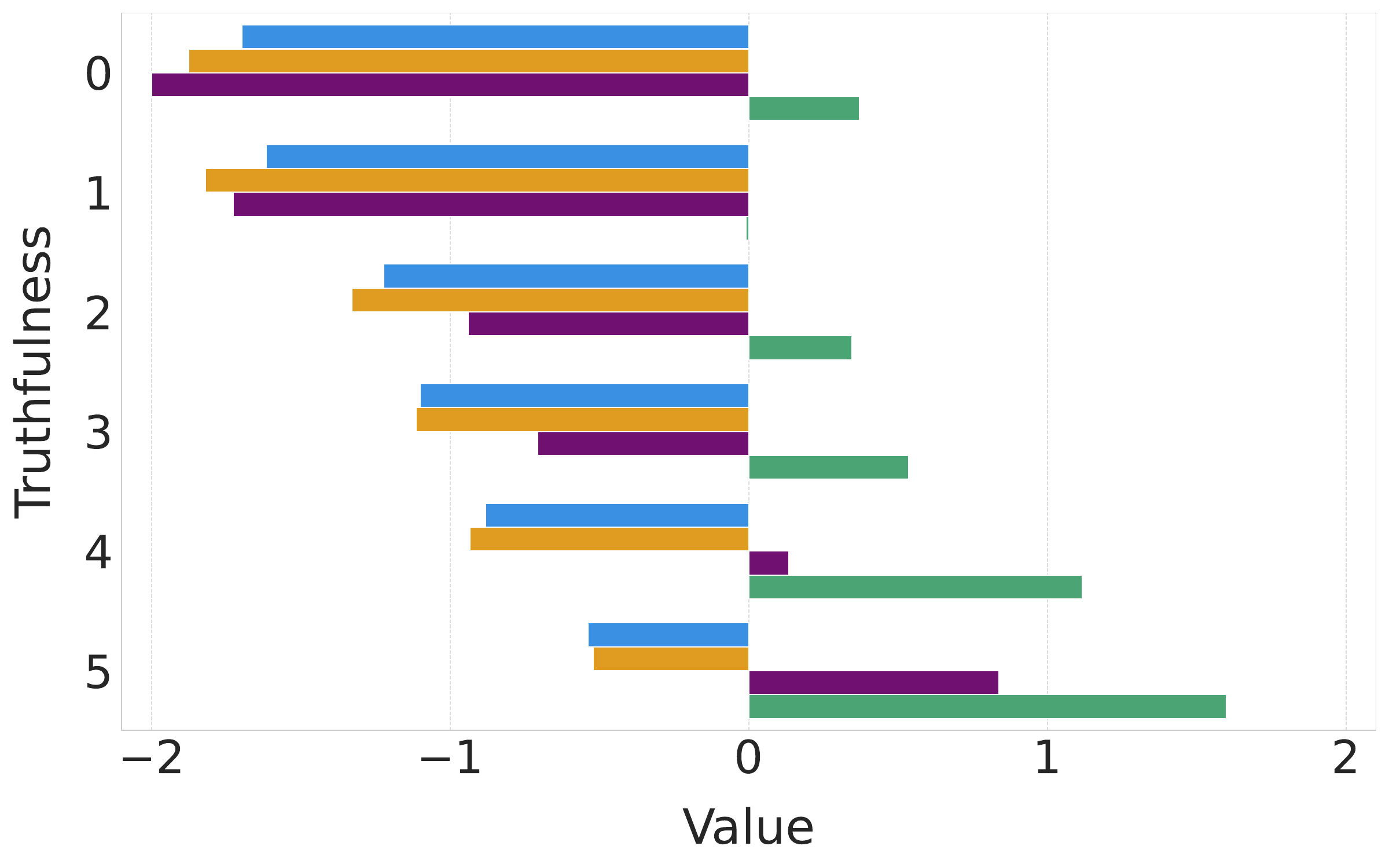}
        \caption{Completeness} \label{fig:Dimensions_Completeness}
    \end{subfigure}
    \begin{subfigure}{0.32\textwidth}
        \centering
        \includegraphics[width=\textwidth]{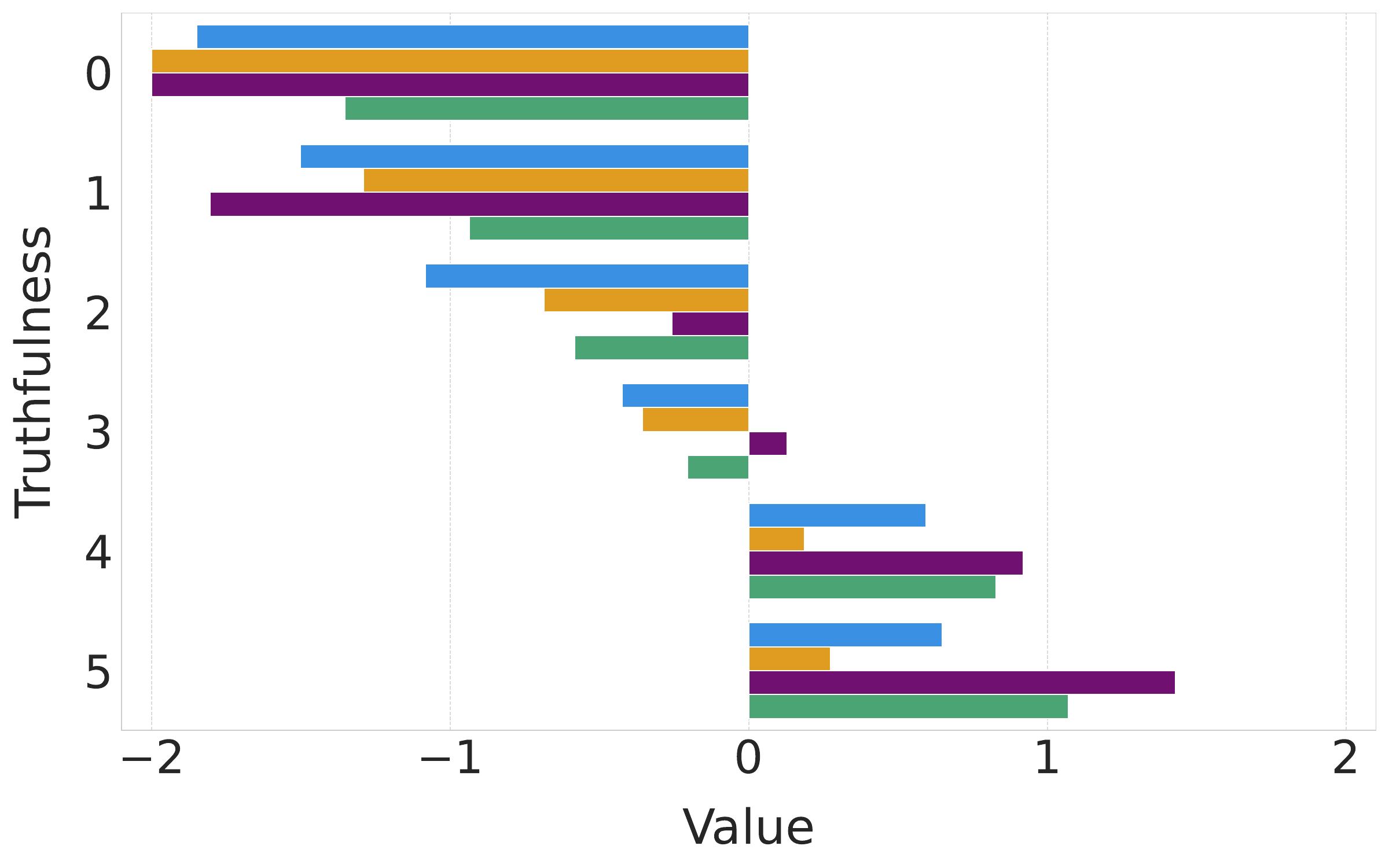}
        \caption{Speaker's Trustworthiness} \label{fig:Dimensions_Trustworthiness}
    \end{subfigure}
    
    \begin{subfigure}{0.32\textwidth}
        \centering
        \includegraphics[width=\textwidth]{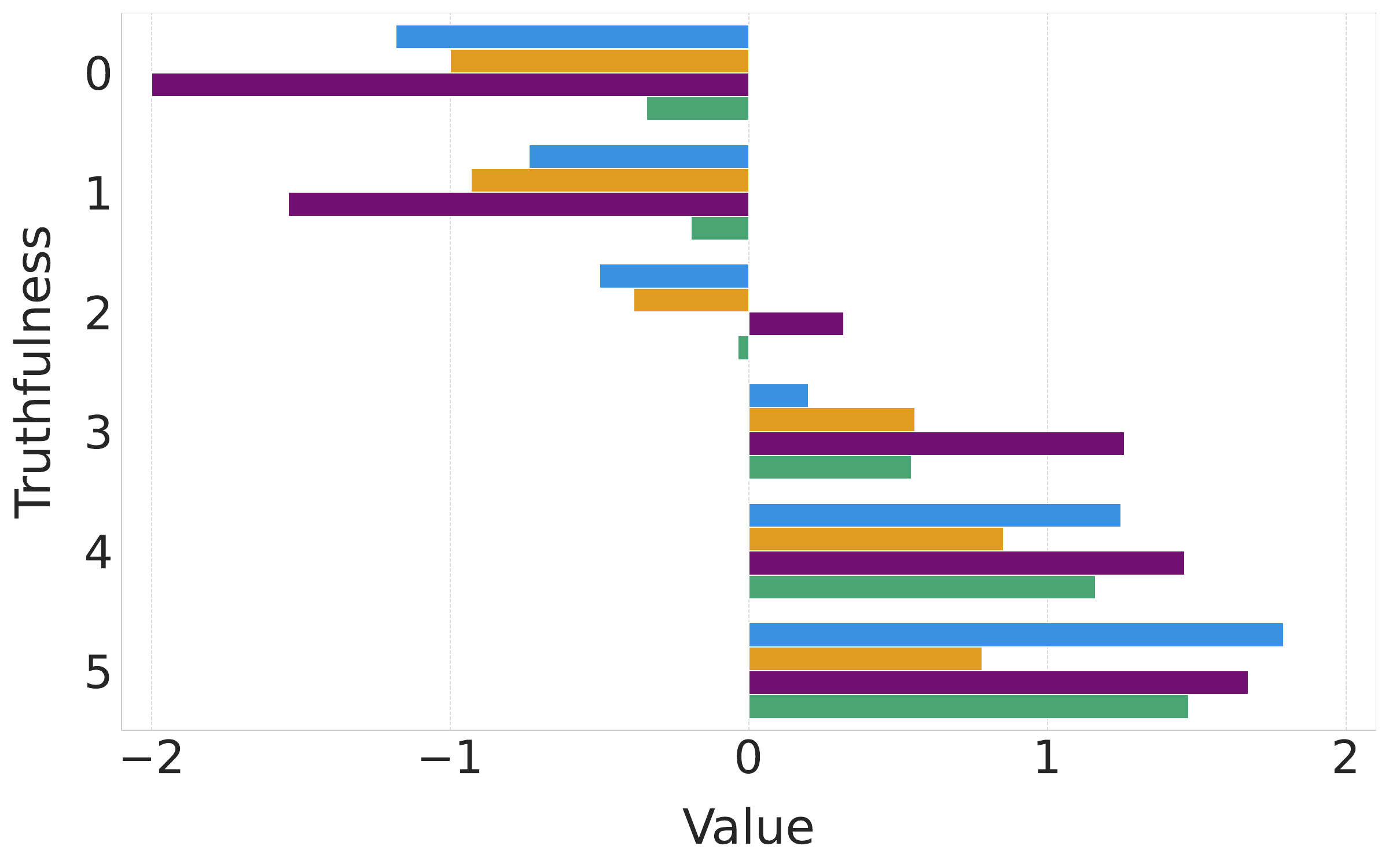}
        \caption{Informativeness} \label{fig:Dimensions_Informativeness}
    \end{subfigure}
    \hspace{0.1cm}
    \begin{subfigure}{0.33\textwidth}
        \includegraphics[width=\textwidth]{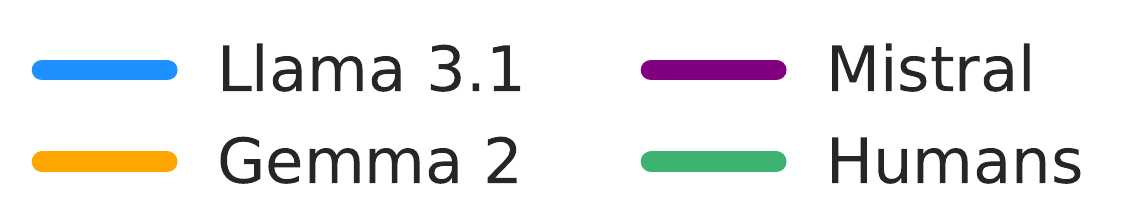}
    \end{subfigure}
    \caption{{\color{black}\textbf{Average ratings assigned by generative agents and human evaluators to each quality dimension across different truthfulness levels.} Generative agents consistently rated \textit{"Comprehensibility"} higher and were more stringent in \textit{"Completeness"}, with ratings rising alongside truthfulness. Human evaluators showed greater variability and gave lower scores for less truthful statements. Furthermore, agents demonstrated greater sensitivity to truthfulness in dimensions like \textit{"Accuracy"} and \textit{"Precision"}. while human ratings were more moderate and less polarized.}}
    \label{fig:Dimensions_Frequency}
\end{figure*}

Figure \ref{fig:Dimensions_Frequency} presents the average scores assigned to each quality dimension across all truthfulness levels by generative agents and human evaluators, revealing several key patterns.

Both generative agents and human evaluators consistently assigned, on average, positive ratings to the \textit{"Comprehensibility"} dimension across all truthfulness levels. However, generative agents generally provided higher scores, reflecting a strong bias toward perceiving statements as clear and readable, regardless of truthfulness level. In contrast, human evaluators displayed greater variability, including negative and neutral ratings, and tended to assign lower scores to statements with lower truthfulness levels.

Generative agents also demonstrated greater caution when assigning positive scores to the \textit{"Completeness"} dimension compared to human evaluators, with their average ratings progressively increasing alongside truthfulness levels. This suggests that agents were more stringent in their assessments, reserving higher ratings for statements they perceived as highly truthful.

For other dimensions, including \emph{Accuracy}, \emph{Unbiasedness}, \textit{Precision}, \emph{Speaker's Trustworthiness}, and \emph{Informativeness}, generative agents exhibited greater sensitivity to truthfulness variations, assigning lower average scores to statements they perceived as less truthful. While human evaluators also showed a positive correlation between these dimensions and truthfulness, their ratings were less extreme, indicating a more balanced and moderate evaluation approach.

These findings underscore the differing evaluation tendencies between generative agents and humans. Generative agents appear to rely on stricter and more polarized criteria. In contrast, human evaluators exhibit greater variability and a preference for neutral or moderate ratings, reflecting potentially broader subjective interpretations.

\paragraph*{Correlation}

Table \ref{tab:7_Dimensions_Correlation} highlights the relationship between truthfulness and the other quality dimensions. Overall, generative agents exhibit stronger correlations with truthfulness across most dimensions compared to human evaluators. Specifically, \textit{Accuracy} demonstrates the highest correlation for both groups, with agents achieving a value of 0.92 compared to 0.71 for humans, indicating that high accuracy ratings are consistently associated with higher truthfulness scores in both groups. Similarly, \textit{Unbiasedness}, \textit{Precision}, \textit{Completeness} and \textit{Informativeness} show significantly higher correlations for agents (0.64, 0.82, 0.53 and 0.85, respectively) than for humans (0.58, 0.40, 0.38 and 0.52). This suggests that agents rely more on these dimensions to assess truthfulness.

Conversely, \textit{Speaker's Trustworthiness} exhibits comparable correlations between agents (0.68) and humans (0.67), indicating that both groups similarly associate trustworthiness with truthfulness. Finally, an exception is observed for \textit{Comprehensibility}, where humans display a slightly higher correlation (0.32) than agents (0.20), suggesting that clarity plays a more influential role in human evaluations.

These findings indicate that generative agents not only align closely with human reasoning in several dimensions but also exhibit more robust and consistent correlations between truthfulness and key evaluative metrics.

\begin{table}[t]
    \centering
    \caption{\textbf{Correlation between truthfulness and quality dimensions for generative agents and human evaluators.} Agents exhibit stronger correlations with most dimensions.}
    \label{tab:7_Dimensions_Correlation}
    \resizebox{0.6\columnwidth}{!}{%
    \begin{tabular}{lcc} 
        \toprule
        \textbf{Dimension} 
        & \multicolumn{2}{c}{\textbf{Correlation Value}} \\ 
        \cmidrule(lr){2-3} & \textbf{Agents} & \textbf{Humans} \\
        \midrule
        Accuracy & \textbf{0.92} & 0.71 \\
        Unbiasedness & \textbf{0.64} & 0.58 \\
        Comprehensibility & 0.20 & \textbf{0.32} \\
        Precision & \textbf{0.82} & 0.40 \\
        Completeness & \textbf{0.53} & 0.38 \\
        Speaker's Trustworthiness & \textbf{0.68} & 0.67 \\
        Informativeness & \textbf{0.85} & 0.52 \\
        \bottomrule
    \end{tabular}
    }
\end{table}


\subsubsection{User-Specific Attributes}

We evaluated the accuracy of generative agents and human participants across various demographic and ideological traits. Table \ref{tab:RQ3_Results} presents the accuracy achieved by each group, highlighting the consistent superiority of generative agents across all demographic categories and classification scales. However, distinct patterns of performance variability emerge between the two groups, revealing key differences in their evaluation dynamics. Generative agents exhibit a high degree of consistency, with minimal variability across traits such as ethnicity, political affiliation, education, gender, and age. This consistency is further supported by the results of the Mann-Whitney U test, which revealed no statistically significant differences across these traits ($p > 0.05$). In contrast, human participants show more pronounced fluctuations. For instance, human evaluations display greater variability across education levels and a more marked sensitivity to age and gender differences, with statistically significant differences observed ($p < 0.05$). This suggests that human evaluations are more influenced by demographic factors compared to generative agents.

These findings suggest that, even with detailed demographic system prompts, generative agents do not effectively simulate the variability observed in human evaluators, highlighting a fundamental limitation in their ability to internalize and act upon subjective demographic contexts. {\color{black}This limitation aligns with an emerging body of literature questioning the extent to which LLM-based agents can faithfully emulate human beliefs, preferences, or biases \cite{lei2024fairmindsim, jiang2024real, tennant2024moral, wang2023aligning, wang2025opera}. While alignment techniques such as reinforcement learning from human feedback (RLHF) have improved LLM performance on instruction-following and value-aligned tasks, replicating nuanced human behavior across demographic dimensions remains an open challenge.} While this can be seen as a shortcoming in terms of mimicking human diversity, this absence may serve as a strength in contexts requiring objectivity and uniformity. The reduced demographic sensitivity of generative agents highlights their potential as impartial tools for tasks like fact-checking, where fairness and consistency are critical.

\begin{table*}[t]
    \centering
    \caption{\textbf{Accuracy of generative agents and human participants across demographic and ideological traits.} Generative agents consistently outperform human participants across all demographic categories, showing minimal variability. In contrast, human evaluations exhibit greater fluctuations, particularly for the 6-level truthfulness scale.}
    \label{tab:RQ3_Results}
    \resizebox{\linewidth}{!}{%
    \begin{tabular}{llcccccccc}
        \toprule
        \textbf{Trait} & \textbf{Category} 
        & \multicolumn{4}{c}{\textbf{2-Level Scale}} & \multicolumn{4}{c}{\textbf{6-Level Scale}} \\
        \cmidrule(lr){3-6} \cmidrule(lr){7-10}
        & & \textbf{Llama 3.1} & \textbf{Gemma 2} & \textbf{Mistral} & \textbf{Humans} 
        & \textbf{Llama 3.1} & \textbf{Gemma 2} & \textbf{Mistral} & \textbf{Humans} \\
        \midrule
        \multirow{2}{*}{\textbf{Ethnicity}} 
        & White & 0.935 & 0.887 & 0.884 & 0.733 & 0.548 & 0.479 & 0.443 & 0.395 \\
        & Black & 0.905 & 0.863 & 0.881 & 0.671 & 0.542 & 0.417 & 0.399 & 0.308 \\
        \midrule
        \multirow{3}{*}{\textbf{Political Faction}}
        & Democrats & 0.905 & 0.890 & 0.875 & 0.736 & 0.530 & 0.470 & 0.438 & 0.420 \\
        & Republicans & 0.952 & 0.863 & 0.857 & 0.683 & 0.554 & 0.470 & 0.440 & 0.320 \\
        & Independents & 0.954 & 0.872 & 0.898 & 0.711 & 0.551 & 0.469 & 0.418 & 0.370 \\
        \midrule
        \multirow{6}{*}{\textbf{Education Level}}
        & Post-graduate Degree & 0.921 & 0.897 & 0.881 & 0.726 & 0.524 & 0.484 & 0.500 & 0.403 \\
        & Post-graduate Schooling & 0.905 & 0.833 & 0.857 & 0.703 & 0.452 & 0.286 & 0.310 & 0.297 \\
        & Bachelor's Degree & 0.921 & 0.854 & 0.875 & 0.721 & 0.543 & 0.454 & 0.429 & 0.383 \\
        & College & 0.929 & 0.948 & 0.890 & 0.703 & 0.584 & 0.526 & 0.429 & 0.401 \\
        & High School & 0.976 & 0.857 & 0.905 & 0.734 & 0.548 & 0.488 & 0.429 & 0.337 \\
        & Less than High School & 1.00 & 0.714 & 0.643 & 0.688 & 0.429 & 0.500 & 0.357 & 0.312 \\
        \midrule
        \multirow{4}{*}{\textbf{Age}}
        & 19-25 & 0.943 & 0.914 & 0.886 & 0.729 & 0.586 & 0.557 & 0.471 & 0.367 \\
        & 26-35 & 0.943 & 0.895 & 0.895 & 0.693 & 0.519 & 0.452 & 0.433 & 0.346 \\
        & 36-50 & 0.933 & 0.861 & 0.893 & 0.727 & 0.560 & 0.433 & 0.425 & 0.389 \\
        & 51-80 & 0.905 & 0.869 & 0.827 & 0.746 & 0.524 & 0.512 & 0.429 & 0.415 \\
        \midrule
        \multirow{2}{*}{\textbf{Gender}}
        & Male & 0.931 & 0.871 & 0.879 & 0.702 & 0.543 & 0.467 & 0.440 & 0.360 \\
        & Female & 0.929 & 0.889 & 0.875 & 0.743 & 0.539 & 0.475 & 0.421 & 0.411 \\
        \bottomrule
    \end{tabular}
    }
\end{table*}

\begin{tcolorbox}[colback=gray!10, colframe=gray!80, title=Takeaway (RQ3)]
Generative agents, similar to humans, prioritize \emph{Accuracy}, \emph{Completeness} and \emph{Speaker's Trustworthiness} for truthfulness assessment. By contrast,  they exhibit lower variability across demographic attributes than human annotators.
\end{tcolorbox}

{\color{black}
\section{Discussion} \label{sec:Discussion}

Our study demonstrates that generative agents, when equipped with demographically grounded profiles and tasked with structured fact-checking procedures, can function as credible proxies for human crowds.  While our results provide detailed evidence of performance, consistency, and bias resilience, their broader implications speak directly to the evolving landscape of misinformation research and mitigation.

From a performance standpoint (RQ1), generative agents consistently produce reliable truthfulness judgments, frequently aligning with expert-verified ground truth and, in some cases, outperforming human crowd. This reliability holds across a wide range of topics and is preserved even as the crowd size varies, demonstrating that high-quality outputs can be achieved with relatively small synthetic groups. Importantly, generative agents exhibit significantly higher internal agreement than human annotators. This consistency not only reflects greater procedural stability but also has substantial downstream value: in real-world settings where fact-checking may be noisy, time-constrained, and subject to interpretive variance, a method that reduces judgment variability can increase trust in the overall system.

These findings suggest that generative agents could help address two of the most persistent challenges in misinformation mitigation: scalability and consistency. Human-led fact-checking cannot keep pace with the volume of online content, and while crowdsourcing can distribute the burden, it introduces substantial variability and bias. Our results show that generative agents may offer a practical middle ground—providing consistent, diverse, and scalable evaluations that approximate crowd judgment while remaining computationally efficient.

At the same time, the mechanisms driving these judgments (RQ2) reveal the limitations and dependencies of current models. While pretrained LLMs contain substantial latent knowledge, our results show that access to external evidence is essential for maintaining accuracy, particularly for recent claims. This highlights the need to design fact-checking pipelines that integrate retrieval components capable of delivering timely and relevant evidence. Conversely, the finding that both humans and agents struggle with very old claims—even in the presence of evidence—highlights the importance of contextualizing claims temporally in automated verification systems and adapting strategies for fact-checking archival content or long-term misinformation narratives.

Our analysis of the evaluative process (RQ3) further illustrates that generative agents apply veracity heuristics similar to those of humans. Both groups prioritize dimensions such as \emph{Accuracy}, \emph{Completeness}, and \emph{Speaker's Trustworthiness}, suggesting that LLM-based agents are capable not only of mimicking surface-level responses but also of emulating meaningful evaluative reasoning. While this alignment supports the validity of using agents as proxies in annotation tasks, subtle differences in the weighting of these dimensions reveal opportunities for greater interpretability in AI-based fact-checking systems. By explicitly modeling how each evaluative trait contributes to a final decision, we can build systems that are not just accurate but also explainable and auditable—crucial features for public accountability and trust.

Lastly, our findings on demographic and ideological variation (RQ3) have direct implications for fairness in misinformation interventions. Whereas human judgments often fluctuate across ideological lines or demographic backgrounds, generative agents exhibited more uniform behavior across simulated profiles. Although this does not imply the absence of bias—agents may still reflect systemic patterns from their training data—it suggests that they are less susceptible to individual-level cognitive or ideological variance. In practice, this opens the door to deploying synthetic crowds as a way to reduce bias amplification in large-scale moderation or truthfulness labeling systems. However, this promise must be tempered with ongoing scrutiny. As discussed in our limitations, the ability of LLMs to authentically model human subjectivity remains an open research area.
}

{\color{black}\subsection{Limitations} \label{sec:Limitations}}

We acknowledge some limitations of our study.

First, our analysis is constrained by the small sample size of fact-checked claims. While our dataset aligns with prior crowdsourced fact-checking research \cite{Crowd_Fact_Checking_Base_Paper}, benchmarking studies typically involve thousands of claims to ensure robust conclusions. The limited number of claims may not fully capture the diversity and complexity of real-world misinformation. Expanding the dataset is crucial to validate the generalizability of our conclusions.

Second, the high agreement observed among generative agents suggests that their fact-checking behavior may be overly consistent, possibly due to prompt formulation or inherent model biases. Future work should explore the effects of prompt variation, different temperature settings, and alternative role-playing approaches to enhance the realism of agent behaviors.

{\color{black}Third, we acknowledge that the ability of LLMs to reproduce human subjectivity-including human demographics-is an open rapidly-evolving research field \cite{lei2024fairmindsim, jiang2024real, tennant2024moral, wang2023aligning, wang2025opera}. Thus, the invariance that we have observed across all demographics attributes may not generalise to new instruction techniques or inherently human-aligned LLMs. 

}

Fourth, while our framework includes an evidence selection phase, this design does not fully reflect the multi-source evaluation typical of human fact-checking workflows. In practice, human evaluators often cross-reference multiple sources to assess a claim’s validity. The reliance on a single pre-selected source simplifies experimental control but may limit the richness of the verification process. Integrating retrieval-augmented generation (RAG) or enabling agents to select and reason over multiple documents would better approximate real-world conditions and allow for more complex reasoning chains.

Finally, the ethical implications of LLM-based fact-checking require deeper consideration. Automated systems raise concerns regarding transparency, accountability, and the potential for misuse. While our study highlights the advantages of generative agents, further research is needed to ensure responsible deployment in real-world fact-checking environments.
\\

\section{Conclusion and Future Work} \label{sec:Conclusion}

This study explores the potential of generative agents as a scalable and effective alternative to human crowds for crowdsourced fact-checking. By leveraging a framework that simulates the crowdsourced fact-checking process using generative agents, we have demonstrated that these agents can address key limitations of existent fact-checking methods, offering a scalable solution to counter misinformation and a significant opportunity to enhance the accuracy of crowdsourced fact-checking systems.

Our findings reveal that generative agents consistently outperform human crowds in fact-checking tasks, achieving superior alignment with expert judgments, especially in scenarios requiring nuanced understanding. They also exhibit greater internal consistency and reduced variability in their assessments. In contrast, human annotators display greater variability and are more susceptible to subjective influences. Furthermore, generative agents maintain impartiality and objectivity, showing a reduced susceptibility to biases. {\color{black}Importantly, agents maintain strong performance even on entirely novel claims published after the LLMs’ training cut-off dates, confirming that their behavior is guided primarily by the evidence they receive rather than memorized knowledge.} These attributes validate their potential as an alternative to human crowds in combating misinformation.

Future work will focus on extending our study by addressing two specific areas derived from our current findings. First, we aim to assess the efficacy of generative agents in domain-specific fact-checking contexts, such as health-related misinformation or financial claims. Additionally, we plan to investigate the ethical implications of deploying generative agents, focusing on transparency, accountability, and the mitigation of potential misuse. We consider pursuing these directions essential to ensure the responsible and widespread adoption of generative agents in real-world fact-checking applications.

\appendix
\section{Prompt Design} \label{Prompt_Design_Appendix}

This appendix provides a comprehensive overview of the structured prompts employed within our framework. Variables enclosed in curly brackets are dynamically substituted throughout the simulation process. A detailed explanation of each variable's role and significance is presented in the corresponding sections.

\subsection{System Prompt} \label{app:System_Prompt}

The System Prompt serves as the foundational instruction set guiding the behavior of the generative agent within our framework. It defines the agent’s role, objectives, and contextual attributes. Each agent profile is assigned based on the real profiles of participants in \cite{Crowd_Fact_Checking_Base_Paper}, ensuring that the demographic and ideological characteristics of the generative agents correspond directly to those of the human participants in the original study. This approach facilitates direct comparison between the results obtained from generative agents and the human crowd, enabling the study of variations in fact-checking behavior under different demographic and ideological conditions.

\begin{tcolorbox}[colback=gray!10, colframe=gray!80, title=System Prompt]
    \footnotesize
    You are a participant in a crowd-based fact-checking program, where your role is to critically evaluate information and statements, and assess its accuracy.
    \\
    
    As a \{age\} year old \{gender\} of \{ethnicity\} ethnicity, you bring a unique perspective to this program.
    \\
    
    You were born in \{birth\_country\} and currently reside in \{residence\_country\}.
    \\
    
    In terms of political alignment, you identify as a \{political\_party\} with generally \{political\_views\} views. 
    \\
    
    Your highest level of education is \{education\_level\}, and your annual family income last year was in the range \{income\_range\}.
    \\
    
    Your views on environmental policies reflect that you \{climate\_change\_stance\} government action to prevent climate change. Regarding the proposal to build a wall along the southern border, you \{border\_wall\_stance\}.
    \\
    
    You are fluent in \{languages\} and you \{student\_status\} a student.
    \\
    
    Your current employment status is \{employment\_status\}.
    \\
    
    Use your background, perspectives, and skills to contribute thoughtfully and objectively to this fact-checking program.
\end{tcolorbox}

Table \ref{tab:system_prompt_variables} details the name of each variable, a brief description of its meaning, and the possible values each attribute can take to define the agent’s identity and perspective.

\begin{table*}[t]
    \centering
    \caption{\textbf{Variables and possible values in the system prompt.} This table outlines the variables used in the system prompt, their descriptions, and the possible values assigned during the simulation.}
    \label{tab:system_prompt_variables}
    \resizebox{\linewidth}{!}{%
    \begin{tabular}{llp{10cm}}
        \toprule
        \textbf{Variable} & \textbf{Description} & \textbf{Possible Values} \\
        \midrule
        \texttt{age} & Age of the agent & 19-25, 26-35, 36-50, 51-80+ \\
        \midrule
        \texttt{gender} & Gender identity & Male, Female \\
        \midrule
        \texttt{ethnicity} & Ethnic background & Asian, Black, White, Mixed \\
        \midrule
        \texttt{birth\_country} & Country of birth & USA, Canada, UK, Germany, France, Other \\
        \midrule
        \texttt{residence\_country} & Current country of residence & USA, Canada, UK, Germany, France, Other \\
        \midrule
        \texttt{political\_party} & Political affiliation & Democrat, Republican, Independent \\
        \midrule
        \texttt{political\_views} & General political orientation & Very Liberal, Liberal, Moderate, Conservative, Very Conservative \\
        \midrule
        \texttt{education\_level} & Highest educational attainment & Less than High School, High School, College, Bachelor's Degree, Post-graduate Schooling, Post-graduate Degree \\
        \midrule
        \texttt{income\_range} & Annual family income & $<20K$, $20K-30K$, $30K-50K$, $50K-100K$, $>100K$ \\
        \midrule
        \texttt{climate\_change\_stance} & View on climate policies & Support, Oppose, Neutral \\
        \midrule
        \texttt{border\_wall\_stance} & Opinion on border wall construction & Support, Oppose, Neutral \\
        \midrule
        \texttt{languages} & Spoken languages & English, Spanish, French, Twi \\
        \midrule
        \texttt{student\_status} & Whether the agent is a student & You are, You are not \\
        \midrule
        \texttt{employment\_status} & Employment situation & Full-time, Part-time, Unemployed, Retired, Student \\
        \bottomrule
    \end{tabular}
    }
\end{table*}

\FloatBarrier

\subsection{Main Prompt} \label{app:Main_Prompt}

The Main Prompt defines the task of the generative agent, which involves verifying the truthfulness of a given statement. The prompt provides a list of relevant web sources, including URLs, titles, and snippets, from which the agent must select the most appropriate evidence to validate the statement. The agent is instructed to evaluate the relevance, credibility, and depth of each source using their background and perspective.

\begin{tcolorbox}[colback=gray!10, colframe=gray!80, title=Main Prompt (1)]
    \footnotesize
    You are a participant in a crowd-based fact-checking program, tasked with verifying the truthfulness of the following statement:
    \\
    
    \{statement\}
    \\

    Below is a collection of URLs, titles, and snippets of web pages that are relevant to the statement:
    \\
    
    \{evidences\_list\}
    \\
\end{tcolorbox}

\begin{tcolorbox}[colback=gray!10, colframe=gray!80, title=Main Prompt (2)]
\footnotesize
    Your goal is to carefully evaluate these sources, considering relevance, credibility, and depth of information, and select a single evidence that is, according to you, the most appropriate to verify the statement.
    \\
    
    As a participant with your unique perspective, use your knowledge and background to inform your selection.
    \\
    
    Provide your response strictly in JSON format with the following fields:
    \\
    \{
    \begin{itemize}
        \item "url": "selected URL",
        \item "title": "title of the selected page",
        \item "snippet": "snippet of the selected page"
    \end{itemize}
    \}
    \\
    
    Each field must be enclosed in double quotes. Ensure that all values are properly escaped if necessary. Do not include additional text, explanations, or comments outside of the JSON object.
    
\end{tcolorbox}

\subsection{Questionnaire Prompt} \label{app:Questionnaire_Prompt}

The Questionnaire Prompt is designed to systematically assess the quality and reliability of a given statement by evaluating it across several key dimensions. The evaluation process incorporates both the statement itself and a summary of the selected article that provides additional context.

\begin{tcolorbox}[colback=gray!10, colframe=gray!80, title=Questionnaire Prompt (1)]
    \footnotesize
    Your task is to assess 8 metrics for the given statement, using both the statement itself and the provided article, which addresses the same topic.
    The statement to evaluate is:
    \\
    
    \{statement\}
    \\
    
    Related Article:
    \\
\end{tcolorbox}

\begin{tcolorbox}[colback=gray!10, colframe=gray!80, title=Questionnaire Prompt (2)]
\footnotesize
    \{summary\}
    \\
    
    Using the article as context to better understand and assess the statement, evaluate each of the following metrics. Provide a numerical score, the associated meaning, and a concise 3-4 sentence explanation (“reason”) for each choice.
    \\
    
    Metrics:
    \begin{enumerate}
        \item ACCURACY: Assess if the statement accurately reflects the topic without errors or incorrect information;
        \item UNBIASEDNESS: Determine if the statement is neutrally and objectively expressed, avoiding subjective or biased language;
        \item COMPREHENSIBILITY: Rate the statement’s clarity and readability, determining if it is easy to understand;
        \item PRECISION: Evaluate whether the information in the statement is specific and detailed rather than vague or ambiguous;
        \item COMPLETENESS: Assess if the statement provides a full, comprehensive view of the topic, rather than only partial information;
        \item SPEAKER'S TRUSTWORTHINESS: Rate the general trustworthiness of the speaker or author, based on reliability and credibility;
        \item INFORMATIVENESS: Judge if the statement provides valuable information or insights, rather than well-known facts or tautologies.
    \end{enumerate}
    
    Values for Metrics 1-7:
    \begin{itemize}
        \item 2 = completely agree;
        \item 1 = partially agree;
        \item 0 = neutral;
        \item -1 = partially disagree;
        \item -2 = completely disagree.
    \end{itemize}

    \begin{enumerate}
        \item TRUTHFULNESS: Assess the statement’s proximity to the truth based on available information, including the article.
    \end{enumerate}
    
    Values for Truthfulness:
    \begin{itemize}
        \item 5 = True (If the statement is accurate and there’s nothing significant missing);
        \item 4 = Mostly-True (If the statement is accurate but needs clarification or additional information);
        \item 3 = Half-True (If the statement is partially accurate but leaves out important details or takes things out of context);
    \end{itemize}
\end{tcolorbox}

\begin{tcolorbox}[colback=gray!10, colframe=gray!80, title=Questionnaire Prompt (3)]
\footnotesize
    \begin{itemize}
        \item 2 = Mostly-False (If the statement contains an element of truth but ignores critical facts that would give a different impression);
        \item 1 = False (If the statement is NOT accurate or is NOT correct);
        \item 0 = Pants-On-Fire (If the statement is NOT accurate or is NOT correct but also makes a ridiculous claim).
    \end{itemize}

    Provide the output strictly in JSON format with the following fields:
    \\\{
    \begin{itemize}
        \item "accuracy\_value": (an integer that can be either -2 or -1 or 0 or 1 or 2);
        \item "accuracy\_meaning": (a string that can only be either "completely agree" or "partially agree" or "neutral" or "partially disagree" or "completely disagree");
        \item "accuracy\_reason": (your justification of the value assigned to the metric of accuracy for the current statement, as a string enclosed in double quotes);
        \item ...
        \item \{the same for other dimensions\}
        \item ...
    \end{itemize}
    \}
    \\
    
    Each field should be filled according to the evaluation criteria above. Please provide accurate justifications based on both the statement and the article content. Remember your response must only be a valid JSON object, do not add anything else.
\end{tcolorbox}

{\color{black}
\section{Example of Web Page Summarization Process} \label{app:summarization_example}

In this section, we provide a concrete example to illustrate how the summarization process described in the \textit{"Summary Generation"} paragraph was implemented. We first present the prompt used to guide the generation of summaries from raw web content. Then, we show how the process was applied to two web pages associated with a specific fact-checked claim. This example clarifies the structure and quality of the summaries provided to the generative agents and exemplifies the criteria used to retain factual relevance and completeness.

\subsection{Prompt Used for Summarization}

The following prompt was employed to generate summaries from raw web page content, ensuring that the output was concise, informative, and focused on information relevant to the claim:

\begin{tcolorbox}[colback=gray!10, colframe=gray!80, title=Web Page Summarization Prompt]
\footnotesize
    You are a model specialized in producing accurate and concise summaries. You are provided with text that may contain unclean characters or irrelevant fragments automatically extracted from web pages using BeautifulSoup—such as advertisements or unrelated metadata. Disregard such elements and focus exclusively on the relevant content.
    \\
    
    Note: The input text may be split into multiple segments.
    \\
    
    Summary Guidelines:

    \begin{itemize}
        \item Language: The summary must be written in ENGLISH.
        \item Accuracy: Ensure the summary faithfully represents the content of the provided text.
        \item Completeness: Include all relevant elements that support or refute the claim. Do not omit significant details.
        \item Factuality: Base your summary solely on the information contained in the input text. Do not rely on assumptions or external knowledge.
        \item Response Format: Return only the summary as your output, with no additional comments or explanations.
    \end{itemize}
    
    Reference Statement: \{claim\}
    
    Text to Summarize: \{web\_page\_text\}
\end{tcolorbox}

\subsection{Claim and Summarized Web Pages}

To illustrate the summarization process, we report here an example consisting of a fact-checked claim and two associated web pages selected as candidate evidence.

\noindent\textbf{Claim:} The leading cause of death among children is a firearm.

\medskip

\noindent\textbf{Web Page 1:} Senate Majority Leader Chuck Schumer’s claim that “the leading cause of death among children is a firearm” is actually MUCH MORE upsetting than you think\footnote{\url{https://www.davideriknelson.com/sbsb/index.php/2022/06/guns-are-now-the-leading-cause-of-death-for-american-kids/}}
\begin{itemize}
\item \textbf{Generated Summary:} Senate Majority Leader Chuck Schumer's statement that firearms are the leading cause of death among children is accurate when 'children' is defined as individuals between the ages of 1 and 19. Data shows that unlike the general population where gun deaths are predominantly suicides (about 66\%), among children, most gun-related deaths are homicides. In America, guns have become the leading cause of death for children, with the majority of these deaths being due to murder, contrasting adult deaths which are more often caused by diseases.
\end{itemize}

\noindent\textbf{Web Page 2:} Firearms overtook auto accidents as the leading cause of death in children\footnote{\url{https://www.npr.org/2022/04/22/1094364930/firearms-leading-cause-of-death-in-children}}
\begin{itemize}
\item \textbf{Generated Summary:} In 2020, firearms became the leading cause of death among children under the age of 19, overtaking motor vehicle accidents, which held that position for the previous 60 years. According to research from the New England Journal of Medicine, overall firearm-related deaths rose by 13.5\% from 2019 to 2020, with deaths among children aged 1 to 19 increasing by nearly 30\%. The study cites a total of 45,222 firearm-related deaths in the U.S. in 2020, with approximately 4,357 being among children. Although the COVID-19 pandemic may have influenced this increase, the specific reasons remain unclear. Unlike adults, where the majority of gun-related deaths are due to suicide, child firearm deaths are primarily homicides. Despite mass shootings receiving significant media attention, they represent a small fraction of gun-related child fatalities, which typically occur in smaller, daily incidents. The disparity between the decrease in motor vehicle deaths and the rise in firearm deaths underscores the lack of similar scientific and policy interventions in addressing firearm violence.
\end{itemize}
}

\bibliographystyle{elsarticle-num}
\bibliography{bibliography}

\end{document}